\documentclass[10pt,journal,compsoc]{IEEEtran}
\ifCLASSOPTIONcompsoc
  \usepackage[nocompress]{cite}
\else
  \usepackage{cite}
\fi
\ifCLASSINFOpdf
\else
\fi

\usepackage{times}
\usepackage{epsfig}
\usepackage{graphicx}
\usepackage{amsmath}
\usepackage{amssymb}
\usepackage{color}
\usepackage{xcolor}
\usepackage{hyperref}
\usepackage{booktabs}
\usepackage{multirow}
\usepackage{xspace}
\usepackage{amsmath}
\usepackage{amssymb}
\usepackage{mathtools}
\usepackage{amsthm}
\usepackage{caption}
\usepackage{url}
\usepackage{algorithm}
\usepackage[noend]{algorithmic}
\usepackage{mdframed}
\usepackage{tcolorbox}
\usepackage{multirow}
\usepackage{tabularx}
\usepackage{float}
\usepackage{graphicx}
\usepackage{color,xcolor}
\usepackage{booktabs}
\usepackage{arydshln}
\usepackage{enumitem}
\usepackage{wrapfig}
\usepackage{caption}
\usepackage{graphicx}
\usepackage{makecell}
\usepackage{float} 
\usepackage{pifont}
\usepackage{tikz}

\usepackage{colortbl}

\usepackage{makecell}

\begin{document}
%
\title{Optimizing Multispectral Object Detection: A Bag of Tricks and Comprehensive Benchmarks}

\author{Chen Zhou$^{\dagger}$ \quad Peng Cheng$^{\dagger}$ \quad Junfeng Fang  \quad Yifan Zhang \quad Yibo Yan \quad Xiaojun Jia \quad Yanyan Xu$^{\ddagger}$ \quad Kun Wang$^{\ddagger}$    \quad Xiaochun Cao \thanks{Yanyan Xu and Kun Wang are corresponding authors. Chen Zhou and Peng Cheng are contributed equally to this work. \\ Chen Zhou, Peng Cheng and Yanyan Xu are with the Beijing Forestry University. \\ Yibo Yan is with the Hong Kong University of Science and Technology. \\  Yifan Zhang is with the State Key Laboratory of Multimodal Artificial Intelligence Systems (MAIS). \\ Junfeng Fang is with the National University of Singapore. \\ Xiaojun Jia and Kun Wang is with the Nanyang Technological University. Xiaochun Cao is with the Sun Yat-sen University.} \\
		{\tt\footnotesize kevinzc9@bjfu.edu.cn } \quad \tt\footnotesize xuyanyan@bjfu.edu.cn \quad \tt\footnotesize wk520529wjh@gmail.com \\ \tt\footnotesize caoxiaochun@mail.sysu.edu.cn  }

\markboth{IEEE TPAMI, Special Issue on Graphs in Vision and Pattern Analysis.}
{Shell \MakeLowercase{\textit{et al.}}: Bare Demo of IEEEtran.cls for Computer Society Journals}
%



\IEEEtitleabstractindextext{%
\begin{abstract}
Multispectral object detection, utilizing RGB and TIR (thermal infrared) modalities, is widely recognized as a challenging task. It requires not only the effective extraction of features from both modalities and robust fusion strategies, but also the ability to address issues such as spectral discrepancies, spatial misalignment, and environmental dependencies between RGB and TIR images. These challenges significantly hinder the generalization of multispectral detection systems across diverse scenarios. Although numerous studies have attempted to overcome these limitations, it remains difficult to clearly distinguish the performance gains of multispectral detection systems from the impact of these ``optimization techniques''. Worse still, despite the rapid emergence of high-performing single-modality detection models, there is still a lack of specialized training techniques that can effectively adapt these models for multispectral detection tasks. The absence of a standardized benchmark with fair and consistent experimental setups also poses a significant barrier to evaluating the effectiveness of new approaches. To this end, we propose the first fair and reproducible benchmark specifically designed to evaluate the training ``techniques'', which systematically classifies existing multispectral object detection methods, investigates their sensitivity to hyper-parameters, and standardizes the core configurations. A comprehensive evaluation is conducted across multiple representative multispectral object detection datasets, utilizing various backbone networks and detection frameworks. Additionally, we introduce an efficient and easily deployable multispectral object detection framework that can seamlessly optimize high-performing single-modality models into dual-modality models, integrating our advanced training techniques. Our codes are available: \url{https://github.com/cpboost/double-co-detr} 
\end{abstract}

\begin{IEEEkeywords}
Multispectral object detection, Multimodal feature fusion, Spatial alignment, Data augmentation
\end{IEEEkeywords}}

\maketitle

\IEEEdisplaynontitleabstractindextext

%
\IEEEpeerreviewmaketitle


\IEEEraisesectionheading{\section{Introduction}\label{sec:introduction}}

%
%
%
%

Multispectral object detection is a powerful technology that leverages both visible light and infrared spectra for object detection, and it has been widely adopted in various real-world applications \cite{jha2021real,kumar2020yolov3,balasubramaniam2022object,carranza2020performance,zuo2021application,qiu2019automatic}, including anomaly detection in surveillance systems \cite{joshi2012survey,mishra2016study,presti2008real,varma2013object,nascimento2006performance}, obstacle recognition in autonomous vehicles \cite{carranza2020performance,nabati2019rrpn,lu2017no,peng2021uncertainty,feng2021review}, defect identification in industrial inspection \cite{zuo2021application,qiu2019automatic,he2019defect,wang2022wafer,yuan2023identification}, and threat detection in defense and security \cite{tribe2004hidden,akcay2022towards,sigman2020background}, to name just few. While many traditional object detection algorithms \cite{zuo2021application,qiu2019automatic,wang2022wafer,liang2019toward,tribe2004hidden} have primarily relied on information from a single modality, recent advancements have explored more sophisticated multispectral architectures \cite{hou2024object,ji2024dual,yan2024iv,feng2020deep,xu2017multi,li2022deepfusion,guo2021deep,xu2022weakly}. In numerous cases, fully exploiting the information from multiple-modalities has demonstrated significant advantages \cite{li2022deepfusion}. For instance, in low-light conditions, leveraging infrared spectra can enhance the performance of visible light detection, or in complex scenarios, combining information from both spectra can improve detection accuracy \cite{geng2019low,yang2019air,wan2023mffnet,ghari2024pedestrian}. Recently, with the rapid development of satellite remote sensing and thermal imaging technologies \cite{wang2022wafer}, many challenging detection datasets have emerged (such as low light and extreme weather conditions) \cite{joshi2012survey,wang2022wafer}. Multispectral detection architectures have demonstrated strong performance on these datasets \cite{qiu2019automatic,wang2022wafer,liang2019toward}.

However, training multispectral object detection models is known to be highly challenging \cite{hou2024object,li2022deepfusion,guo2021deep,wang2019quality,jiang2020cmsalgan,song2022novel}. Beyond the common issues encountered in training deep architectures, such as vanishing gradients and overfitting \cite{liang2019toward,li2022deepfusion}, multispectral models face several \textbf{unique} challenges that limit their strides on these datasets:

\begin{itemize}[leftmargin=*]
    \item[\ding{228}] The first challenge lies in \textit{effectively utilizing dual-modality data}. Simultaneously processing visible and infrared data increases the complexity of dual-modality feature fusion, which may result in suboptimal integration of information from both modalities \cite{hou2024object,jiang2020cmsalgan}. This issue is particularly pronounced in earlier multispectral models, where the fusion process often led to information loss, preventing the models from fully leveraging the strengths of both modalities \cite{wang2019quality,jiang2020cmsalgan}. Additionally, registration discrepancies between the two modalities and the lack of modality-specific enhancement strategies further constrain model performance \cite{song2022novel}.

    \item[\ding{228}] The second major question is the \textit{lack of an effective optimization strategy} for converting high-performance single-modality models into dual-modality models. Despite the emergence of numerous powerful single-modality object detection frameworks in recent years \cite{takken2024hardware,diwan2023object,fang2019tinier,huang2018yolo}, there has yet to be a robust method for effectively harnessing the potential of these models while addressing the unique challenges of multispectral object detection. 
\end{itemize}

To addess the aforementioned challenges, the promising approaches can be categorized into 
\ding{192} dual-modality architectural fusion \cite{feng2020deep,xu2017multi,li2022deepfusion} and \ding{193} modality-specific enhancements \cite{geng2019low,yang2019air,wan2023mffnet}, both of which we classify as ``training techniques''. The former involves adapting single-modality architectures to dual-modality structures, integrating advanced backbone networks, and employing diverse feature fusion strategies. The latter focuses on processing data from both modalities using techniques such as modality-specific data augmentation and alignment calibration \cite{qiu2019automatic}. While these techniques generally contribute to the effective training of multispectral object detection models, their benefits are not always significant or consistent \cite{wang2019quality,jiang2020cmsalgan,song2022novel}. Furthermore, it is often difficult to distinguish the performance improvements achieved through more complex dual-modality architectures from those gained via these ``training techniques''.

In some extreme cases, contrary to initial expectations, single-modality models enhanced with certain optimization techniques may even outperform carefully designed, complex dual-modality architectures \cite{xu2017multi,li2022deepfusion,guo2021deep,xu2022weakly}. This casts doubt on the pursuit of increased complexity, thereby rendering it a less attractive approach. \textbf{These observations highlight a critical gap in the study of multispectral object detection: the lack of a standardized benchmark that can fairly and consistently evaluate the effectiveness of training techniques for dual-modality models.} Without disentangling the effects of architectural complexity from the ``training techniques'' applied, it may remain unclear whether multispectral object detection should inherently perform better under otherwise identical conditions.

\vspace{-2em}
\subsection*{Our Contribution}
To establish such a fair benchmark, our first step was to conduct a comprehensive investigation into the design philosophies and implementation details of dozens of popular multispectral object detection techniques, including various backbone networks, dual-modality fusion strategies, and alignment techniques. Unfortunately, we discovered that even on the same datasets, the implementation of hyperparameter configurations (such as hidden layer dimensions, learning rates, weight decay, dropout rates, number of training epochs, and early stopping patience) is highly inconsistent and often varies depending on specific circumstances. This inconsistency makes it challenging to draw any fair or reliable conclusions.

To this end, we conducted a detailed analysis of these sensitive hyperparameters and standardized them into a ``best'' hyperparameter set, consistently applied across all experiments. This standardization provides a fair and reproducible benchmark for training multispectral object detection models. Subsequently, we explored various combinations of training techniques across several classical multispectral object detection datasets, leveraging common single-modality model backbones and optimizing them for dual-modality detection tasks.

The results of our comprehensive study were highly significant. Based on the characteristics of different single-modality model backbones, framework features, and detection sample characteristics, we developed several effective training techniques and optimization strategies, enabling us to achieve state-of-the-art results on multiple representative datasets. Furthermore, we proposed several optimization strategies with strong transferability, demonstrating excellent performance across multiple dual-modality public datasets \footnote{Our research was awarded the championship in the Global Artificial Intelligence Innovation Competition (GAIIC) \url{https://gaiic.caai.cn/ai2024}, out of over 1,200 participants, 1,000+ teams, and 8,200+ submissions.}.

Specifically, our contributions are as follows:

\ding{229} \textbf{Multimodal Feature Fusion:} we introduce advanced multimodal feature fusion techniques to effectively integrate visible and infrared data, enhancing the feature representation capabilities of multispectral object detection models, especially in complex environments.

\ding{229} \textbf{Dual-Modality Data Augmentation:} we employ modality-specific data augmentation strategies that cater to the distinct characteristics of visible and infrared data, improving the model's robustness in varying environmental and complex scenarios.

\ding{229} \textbf{Alignment Optimization:} by implementing precise alignment techniques, we improve spatial consistency between visible and infrared data, reducing inter-modality misalignment and significantly enhancing performance in low-light object detection and multimodal information fusion.

\ding{229} \textbf{Optimizing Single-Modality Models for Dual-Modality Tasks:} we provide a new benchmark and training techniques to effectively adapt high-performing single-modality models into dual-modality detection models. Through these optimizations, single-modality models outperform even complex, large-scale dual-modality detection models, offering strong support for their migration to dual-modality tasks.






\section{Related work}

\subsection{Multispectral Object Detection \& Training Challenges}
Multispectral object detection has achieved state-of-the-art performance in applications like autonomous driving and drone-based remote sensing \cite{zuo2021application,qiu2019automatic,joshi2012survey,mishra2016study,presti2008real,varma2013object}. However, implementing multispectral detection is challenging, especially when dealing with images from distinct spectra, such as visible light (RGB) and thermal infrared (TIR) \cite{nascimento2006performance,he2019defect,wang2022wafer}. Existing methods \cite{wang2022bimodal,yan2020bimodal,zhang2017discriminative,zheng2023inspection} face several issues, including spectral differences \cite{tang2024revisiting}, spatial misalignment \cite{lu2024after}, and high sensitivity to environmental conditions \cite{cao2024anchor}, limiting their generalization across diverse scenarios. While recent studies have introduced various training techniques, they often struggle to deliver consistent performance improvements when applied to complex remote sensing data \cite{dai2021tirnet,tang2023exploring}, differing from the dual-modality detection benchmarks discussed in this paper.

To address these challenges, techniques such as multimodal feature fusion, registration alignment, and dual-modality data augmentation have been developed in recent years \cite{ji2024dual,yan2024iv,feng2020deep,xu2017multi,li2022deepfusion}. The following sections provide a detailed exploration of these techniques and their applications.

\subsection{Multimodal Feature Fusion}
In multispectral object detection, feature fusion plays a crucial role in enhancing model performance. Current fusion methods are generally categorized into three types: pixel-level, feature-level, and decision-level fusion. Pixel-level fusion \cite{zhang2019multi,zhang2024exploring,zhu2021rgbt} integrates RGB and TIR images at the input stage, allowing early information combination but potentially introducing noise or misalignment due to differences in resolution and viewpoints. Feature-level fusion \cite{zhu2020quality,tu2022rgbt,zhai2024rgbt,zhou2022defnet,gao2021unified} combines high-level features from both modalities at intermediate layers, utilizing techniques like concatenation, weighting, or attention mechanisms to better capture complementary information, though it may add computational overhead \cite{xiao2023multi,cai2023multi}. Decision-level fusion \cite{tang2023exploring,peng2022dynamic,zhao2023rmfnet,oh2017object,zhang2023illumination,sun2021loftr} merges independent detection results from each modality at the final stage, providing efficiency and stable performance, especially when the modalities offer relatively independent information.

\subsection{Dual-Modality Data Augmentation}

In multispectral object detection, data augmentation is crucial for improving model generalization and reducing overfitting \cite{xu2017multi,li2022deepfusion}. While traditional techniques like flipping, rotation, and scaling work well in single-modality detection \cite{yan2024iv,feng2020deep}, the fusion of RGB and TIR images introduces higher complexity. A common approach is to apply synchronized augmentation to both RGB and TIR images \cite{balasubramaniam2022object,qiu2019automatic} to ensure consistency between the modalities. Techniques such as random cropping, scaling, and color transformations increase image diversity and help the model adapt to varying environmental conditions \cite{orlosky2017vismerge,andersen2024learning}. Additionally, some studies propose joint data augmentation methods, such as mixed modal augmentation, which exchanges pixels or features between modalities to enhance robustness against modality differences \cite{carranza2020performance,zuo2021application,chen2021data,lambrecht2019towards}, ultimately improving detection performance in challenging scenarios.

\subsection{Registration Alignment}
Registration alignment techniques are employed to address spatial discrepancies between images from different sensors, such as RGB and TIR. Differences in resolution and viewpoints often lead to misalignment and distortion, which can negatively impact feature fusion and detection performance \cite{tu2022weakly,yuan2022translation}. Traditional alignment methods \cite{zhang2024amnet,liu2024rgbt,li2024deep}, such as scaling, rotation, and affine transformation, are used to align the images but tend to be limited in complex scenes or when nonlinear deformations are present. Recently, deep learning-based alignment techniques \cite{zhang2023illumination,tang2019rgbt,velesaca2024multimodal,brenner2023rgb,zhang2023drone,tang2022superfusion} have emerged, achieving pixel-level precision by learning feature mappings between RGB and TIR images, and using contrastive loss or self-supervised learning to ensure spatial consistency. Some methods also incorporate attention mechanisms to dynamically adjust feature alignment \cite{lu2024after,tu2022weakly,li2024deep,liu2023quality}, enhancing both local detail and global consistency.
\section{Method}

In this section, we systematically discuss how to improve existing dual-modality object detection algorithms. The discussion focuses on three key aspects: \textbf{multimodal feature fusion, dual-modality data augmentation, and registration alignment.} Specifically, Section 3.1 details the hyperparameter configurations and datasets used in our experiments, while Sections 3.2, 3.3, and 3.4 discuss multimodal feature fusion, dual-modality data augmentation, and registration alignment, respectively.

\begin{table*}[htbp] \footnotesize
\centering
\renewcommand{\arraystretch}{1} 
\setlength{\tabcolsep}{12pt} 
\captionsetup{justification=centering, labelsep=period, font=bf}
\caption{\footnotesize{Configurations of the optimal hyperparameters adopted to implement different single models for training on the KAIST dataset.}} 
\label{tab:training_settings}
\begin{tabular}{lcccc}
\toprule
\textbf{Method} & \textbf{Total epoch} & \textbf{Learning rate \& Decay} & \textbf{Weight decay} & \textbf{Dropout} \\ 
\midrule
YOLOv3 \cite{kumar2020yolov3}          & 100                  & $1 \times 10^{-2}$           & $5 \times 10^{-4}$ & 0.5              \\ 
Faster R-CNN \cite{ren2016fasterrcnnrealtimeobject}    & 80                   & $1 \times 10^{-2}$           & $1 \times 10^{-4}$ & 0.5              \\ 
SSD \cite{Liu_2016}             & 120                  & $2 \times 10^{-3}$           & $5 \times 10^{-4}$ & 0.5              \\ 
RetinaNet \cite{8237586}       & 100                  & $1 \times 10^{-3}$           & $1 \times 10^{-4}$ & 0.3              \\ 
EfficientDet \cite{tan2020efficientdetscalableefficientobject}    & 150                  & $5 \times 10^{-4}$           & $4 \times 10^{-5}$ & 0.3              \\ 
Mask R-CNN \cite{10055028}      & 90                   & $2 \times 10^{-2}$           & $1 \times 10^{-4}$ & 0.5              \\ 
YOLOv5 \cite{geetha2024comparingyolov5variantsvehicle}          & 300                  & $1 \times 10^{-2}$           & $5 \times 10^{-4}$ & 0.5              \\ 
CenterNet \cite{tian2019fcosfullyconvolutionalonestage}        & 140                  & $1 \times 10^{-3}$           & $1 \times 10^{-4}$ & 0.4              \\ 
FCOS \cite{zhou2019objectspoints}            & 120                  & $2.5 \times 10^{-3}$           & $1 \times 10^{-4}$ & 0.5              \\ 
Cascade R-CNN \cite{10055028}   & 100                  & $5 \times 10^{-3}$           & $1 \times 10^{-4}$ & 0.5              \\ 
\bottomrule
\end{tabular}
\end{table*}
\subsection{Standardized Experimental Configuration}

We conducted a comprehensive analysis of previous single-modality object detection models applied to dual-modality detection tasks. To further enhance model performance and improve the robustness of our benchmarking, we also performed hyperparameter optimization and fine-tuning on these models. Based on dual-modality datasets (including both RGB and TIR data), we systematically explored the adaptability of single-modality models in integrating multimodal information, with particular emphasis on their performance across different modalities. The key hyperparameter configurations are presented in Table \ref{tab:training_settings}.

Through a grid search approach, we optimized the hyperparameters for all methods and identified the most generalizable and effective configuration. This configuration was selected based on the best performance of various single-modality detection models across multiple datasets, focusing on parameters such as learning rate, weight decay, and dropout. Ultimately, we proposed this ``optimal hyperparameter configuration'' and strictly adhered to it in our experiments.

\textbf{Specifically, the final configuration consists of a learning rate of 0.01 with decay, a weight decay of 0.0001, and a dropout rate of 0.5.} We believe that this setup provides stable and efficient performance across a range of multispectral object detection tasks, ensuring fair comparisons between different methods under the same conditions.

In each experiment, we trained for up to 200 epochs, with early stopping set to a patience of 20 epochs. To minimize the impact of random variations, each experiment was repeated 20 times, and the results were averaged to obtain the final performance metrics.

Our subsequent experiments utilized the KAIST \cite{hwang2015multispectral}, FLIR, and DroneVehicle \cite{sun2021dronebasedrgbinfraredcrossmodalityvehicle} datasets. The KAIST dataset is a benchmark for pedestrian detection, combining visible and infrared images to evaluate multispectral detection algorithms. The FLIR dataset includes multi-class vehicle and pedestrian detection tasks with high-resolution thermal imagery. The DroneVehicle dataset focuses on multi-class object detection from a drone’s perspective, covering various complex scenarios.

Regarding performance evaluation, we employed task-specific metrics tailored to the characteristics of each dataset. For the KAIST dataset, we selected Miss Rate as the primary evaluation metric due to its sensitivity to missed detections, which is critical in this context. In contrast, for the FLIR and DroneVehicle datasets, we used mean Average Precision (mAP) as the evaluation metric, as these datasets involve multi-class object detection, and mAP provides a more comprehensive assessment of detection accuracy across classes. This dataset-specific approach ensures a thorough and accurate evaluation of each method's performance in diverse tasks.

By leveraging a unified hyperparameter configuration and dataset-specific evaluation metrics, we ensure fair and consistent comparisons between different methods, providing a robust foundation for subsequent performance improvements.

\subsection{Multimodal Feature Fusion}
\textbf{Formulations.} 
In multispectral object detection, multimodal feature fusion techniques aim to effectively integrate complementary information from both RGB and TIR images, enhancing detection accuracy and model robustness. \textbf{Multimodal feature fusion can be categorized into three main approaches: pixel-level fusion, feature-level fusion, and decision-level fusion.}

\begin{enumerate}[leftmargin=*]
\item[\ding{115}] \textbf{Pixel-Level Fusion.} The fusion of RGB and TIR images at the pixel level involves a series of tensor-based transformations, incorporating both adaptive weighting and convolutional refinement to effectively integrate the complementary modalities. Let $\mathcal{I}_{\text{R}} \in \mathbb{R}^{h \times w \times 3}$ denote the RGB image and $\mathcal{I}_{\text{T}} \in \mathbb{R}^{h \times w \times 1}$ denote the TIR image. Initially, the TIR image is expanded to a three-channel format:
\begin{equation}
    \mathcal{I}_{\text{T}}^{'}
 = \mathcal{E}\left(\mathcal{I}_{\text{T}}, \mathcal{J}_{\text{3}}\right) = \sum_{c=1}^{3} \left(\mathcal{I}_{\text{T}} \otimes \mathcal{J}_{3}^{(c)}\right) \in \mathbb{R}^{h \times w \times 3},
\end{equation}
where $\mathcal{J}_{3}$ denotes a tensor of shape \small{$1 \times 1 \times 3$}, employed to replicate the TIR image along the channel dimension via the tensor outer product operation $\otimes$. The summation term $\sum_{c=1}^{3}$ indicates that this operation is applied independently to each channel, expanding the single-channel TIR image $\mathcal{I}_{\text{T}}$ into a three-channel representation $\mathcal{I}_{\text{T}}^{'}$ consistent with the structure of RGB images.

Next, adaptive pixel-wise weighting matrices $\mathcal{W}_{\text{R}} \in \mathbb{R}^{h \times w \times 3}$ and $\mathcal{W}_{\text{T}} \in \mathbb{R}^{h \times w \times 3}$ are introduced for the RGB and TIR channels. The intermediate fused image $\mathcal{I}_{\text{interim}}$ is defined as:
\begin{equation}
    \mathcal{I}_{\text{interim}} = \mathcal{W}_{\text{R}} \odot \mathcal{I}_{\text{R}} + \mathcal{W}_{\text{T}} \odot \mathcal{I}_{\text{T}}^{'} + \boldsymbol{\eta}(\mathbf{n}),
\end{equation}
where $\boldsymbol{\eta}(\mathbf{n})$ represents a noise model parameterized by a Gaussian random vector $\mathbf{n} \sim \mathcal{N}(\mathbf{0}, \boldsymbol{\Sigma})$, accounting for the inherent uncertainty in sensor measurements.

To refine the fusion process and incorporate spatial context, convolutional transformations are applied using modality-specific kernels $\mathcal{K}_{\text{R}} \in \mathbb{R}^{k \times k \times 3 \times 3}$ and $\mathcal{K}_{\text{T}} \in \mathbb{R}^{k \times k \times 3 \times 3}$. The convolutional outputs are expressed as:
\begin{equation}
\varphi_{\ell} = 
\begin{cases}
    \mathcal{K}_{\text{R}} \ast \mathcal{I}_{\text{R}} + \beta_{\text{R}}, & \text{if } \ell = \text{R} \\
    \mathcal{K}_{\text{T}} \ast \Psi_{\text{T}} + \beta_{\text{T}}, & \text{if } \ell = \text{T}
\end{cases},
\end{equation}

where $\ast$ denotes the convolution operation, and $\mathbf{\beta}_{\text{R}}$ and $\mathbf{\beta}_{\text{T}}$ are the learnable bias terms.

The final fused image $\mathcal{I}_{\text{fuse}}$ is obtained through a non-linear fusion strategy that incorporates spatially adaptive weight mappings. We introduce non-linear mapping functions $\mathcal{F}_{\text{R}}(\cdot, \cdot)$ and $\mathcal{F}_{\text{T}}(\cdot, \cdot)$. The  $\mathcal{I}_{\text{fuse}}$ is expressed as:
\begin{equation}
    \mathcal{I}_{\text{fuse}} = \mathcal{F}_{\text{R}}\left(\mathcal{I}_{\text{interim}}, \mathcal{W}_{\text{R}}\right) \odot \varphi_{\text{R}} + \mathcal{F}_{\text{T}}\left(\mathcal{I}_{\text{interim}}, \mathcal{W}_{\text{T}}\right) \odot \varphi_{\text{T}}.
\end{equation}

The functions $\mathcal{F}_{\text{R}}(\cdot, \cdot)$ and $\mathcal{F}_{\text{T}}(\cdot, \cdot)$ can be simplified as follows:
\begin{equation}
\mathcal{F}_{\ell}\left(\mathcal{I}_{\text{interim}}, \mathcal{W}_{\ell}\right) = 
\sigma\left(\mathcal{W}_{\ell} + \boldsymbol{\alpha}_{\ell} 
\odot \delta\left(\mathcal{G}_{\ell}\left(\mathcal{I}_{\text{interim}}\right)\right)\right),
\end{equation}
where $\ell \in \{\text{R}, \text{T}\}$,  $\sigma(\cdot)$ denotes the sigmoid activation function, $\delta(\cdot)$ represents the hyperbolic tangent function, and $\mathcal{G}_{\ell}(\cdot)$ is a non-linear spatial filtering operation. The term $\boldsymbol{\alpha}_{\ell} \in \mathbb{R}^{h \times w \times 3}$ is a learnable scaling tensor.

The proposed pixel-level fusion scheme integrates adaptive weighting, convolutional refinement, and a multi-layered non-linear transformation pipeline to enhance representation capacity. The noise modeling term $\eta(n)$ improves robustness, while the activation functions $\sigma(\cdot)$ and $\delta(\cdot)$ facilitate non-linear interactions between the RGB and TIR modalities.

\item[\ding{115}] \textbf{Feature-Level Fusion.} 
The mainstream feature-level fusion methods primarily include convolution-based Network-in-Network (NIN) modules and bidirectional attention-based Iterative Cross-modal Feature Enhancement (ICFE) modules \cite{shen2023icafusioniterativecrossattentionguided}. The following sections provide an in-depth introduction to each approach.

\textbf{NIN Module}.  
To achieve independent localized non-linear transformations on the RGB and TIR modalities, we first designed a network-in-network module integrated with a residual structure. This module serves as a foundational step for subsequent cross-modal feature enhancement and interaction, where we leverage 1x1 convolutions to apply fine-grained, spatially localized non-linear mappings that improve the expressiveness of feature representations. Let $\mathcal{X}_{\text{R}}^l \in \mathbb{R}^{H \times W \times C}$ and $\mathcal{X}_{\text{T}}^l \in \mathbb{R}^{H \times W \times C}$ denote the RGB and TIR modality feature maps at layer $l$, respectively. We define learnable 1x1 convolution kernels $\mathcal{W}_{\text{R}}^{(1 \times 1)} \in \mathbb{R}^{C \times C}$ and $\mathcal{W}_{\text{T}}^{(1 \times 1)} \in \mathbb{R}^{C \times C}$, applying them with residual connections to each modality for localized feature transformations, as follows:

\begin{equation}
\mathcal{D}_{\ell}^l = 
\begin{cases} 
\mathcal{X}_{\text{R}}^l + \left(\mathcal{W}_{\text{R}}^{(1 \times 1)} \ast \mathcal{X}_{\text{R}}^l + \zeta_{\text{R}}^l\right), & \text{if } \ell = \text{R} \\ 
\mathcal{X}_{\text{T}}^l + \left(\mathcal{W}_{\text{T}}^{(1 \times 1)} \ast \mathcal{X}_{\text{T}}^l + \zeta_{\text{T}}^l\right), & \text{if } \ell = \text{T}
\end{cases}.
\end{equation}

The residual connection within this transformation preserves original modality-specific information in the transformed feature, mitigating potential information loss or distortion. To achieve adaptive fusion of RGB and TIR modalities, we introduce dynamic weighting coefficients \(\alpha_{\text{R}}\) and \(\alpha_{\text{T}}\), computed through a transformation \(\nu(\cdot)\) followed by a shared non-linear function \(\sigma(\cdot)\) applied to each transformed feature map:

\begin{equation}
\alpha_{\text{R}} = \sigma(\nu(\mathcal{D}_{\text{R}}^l)), \quad \alpha_{\text{T}} = \sigma(\nu(\mathcal{D}_{\text{T}}^l)).
\end{equation}

The final fused feature $\mathcal{D}_{\text{fuse}}^l$ at layer $l$ is given as follows:

\begin{equation}
\mathcal{D}_{\text{fuse}}^l = \alpha_{\text{R}} \odot \mathcal{D}_{\text{R}}^l + \alpha_{\text{T}} \odot \mathcal{D}_{\text{T}}^l.
\end{equation}

Through these operations, the NIN module not only performs modality-specific, localized feature transformations but also enables adaptive and balanced feature fusion. This module strengthens the feature discriminability and robustness, while preserving localized information via non-linear activation and residual connections. 

\textbf{ICFE Module}.  
The ICFE module progressively enhances feature representations of RGB and TIR modalities by iteratively exchanging and refining complementary information, ultimately producing a single fused feature representation. Let \(\mathcal{T}_\text{R}^{(0)}\) and \(\mathcal{T}_\text{T}^{(0)}\) represent the initial RGB and TIR features, respectively, and let the final fused feature representation after \(n\) iterations be denoted as $\mathcal{V}_{\text{fuse}}^{(\text{n})}$. The following outlines the detailed formulae of this process.

At the k-th iteration, multi-head queries, keys, and values are generated for both the RGB and TIR modalities. Suppose there are H attention heads, indexed by h. For the h-th attention head, we compute the query matrix \(\mathcal{Q}_\text{R}^{(\text{k,h})}\) for RGB features, and the key matrix \(\mathcal{K}_\text{T}^{(\text{k,h)}}\) and value matrix \(\mathcal{V}_\text{T}^{\text{(k,h})}\) for TIR features:

{\scriptsize
\begin{equation}
\mathcal{Q}_\text{R}^\text{(k,h)} = \mathcal{T}_\text{R}^\text{(k)} \mathcal{W}_\text{Q}^\text{(h)}, \mathcal{K}_\text{T}^\text{(k,h)} = \mathcal{T}_\text{T}^\text{(k)} \mathcal{W}_\text{K}^\text{(h)}, \mathcal{V}_\text{T}^\text{(k,h)} = \mathcal{T}_\text{T}^\text{(k)} \mathcal{W}_\text{V}^\text{(h)},
\end{equation}
}

where \(\mathcal{W}_\text{Q}^\text{{(h)}}, \mathcal{W}_\text{K}^\text{{(h)}}, \mathcal{W}_\text{V}^\text{{(h)}} \in \mathbb{R}^{d \times d_H}\) are learnable projection matrices, and \(\text{d}_\text{H} = \text{d} / \text{H}\) represents the dimensionality per attention head.

To obtain the cross-modally enhanced RGB features \(\mathcal{Z}_\text{R}^{\text{(k,h)}}\), we calculate the weighted matrix by applying the softmax function to the scaled dot product of the query and key matrices, then multiply it with the value matrix:

\begin{equation}
\mathcal{Z}_\text{R}^{\text{(k,h)}} = \textit{softmax}\left(\frac{\mathcal{Q}_\text{R}^{\text{(k,h)}} (\mathcal{K}_\text{T}^{\text{(k,h)}})^\text{T}}{\sqrt{\text{d}_\text{H}}}\right) \mathcal{V}_\text{T}^{\text{(k,h)}}.
\end{equation}

Then, we concatenate the features from all attention heads (denoted by \(\Gamma\) as the concatenation operation) and project them back to the original feature space using an output projection matrix \(\mathcal{W}_\text{O}\):

\begin{equation}
\mathcal{Z}_\text{R}^{(\text{k})} = \Gamma(\mathcal{Z}_\text{R}^{\text{(k,1)}}, \ldots, \mathcal{Z}_\text{R}^{\text{(k,H)}}) \mathcal{W}_\text{O},
\end{equation}
where \(\Gamma(\cdot)\) represents the concatenation operation applied across all attention heads.

In each iteration, the RGB and TIR features are combined to produce an intermediate fused feature representation \(\mathcal{V}_{\text{fuse}
}^{(\text{k})}\), with learnable weighting coefficients \(\lambda^{(\text{k})}\) and \(\mu^{(\text{k})}\) controlling the fusion:

\begin{equation}
\mathcal{V}_{\text{fuse}}^{(\text{k})} = \lambda^{(\text{k})} \odot \mathcal{Z}_\text{R}^{(\text{k})} + \mu^{(\text{k})} \odot \mathcal{Z}_\text{T}^{(\text{k})},
\end{equation}
where \(\mathcal{Z}_\text{T}^{(\text{k})}\) is the cross-modally enhanced TIR feature obtained symmetrically to \(\mathcal{Z}_\text{R}^{(\text{k})}\).

To further enhance non-linear representation capabilities, a non-linear activation function \(\delta(\cdot)\) is applied with residual connection to the fused feature in each iteration. After \(n\) iterations, the final fused feature representation is given by:

\begin{equation}
\mathcal{V}_{\text{fuse}}^{(\text{n})} = \mathcal{V}_{\text{fuse}}^{(\text{n-1})} + \delta\left(\mathcal{V}_{\text{fuse}}^{(\text{n-1})}\right).
\end{equation}

\item[\ding{115}] \textbf{Decision-Level Fusion.} 
In decision-level fusion, RGB and TIR modalities undergo separate feature extraction and preliminary detection, and their fusion occurs at the final decision stage. Let the detection results for RGB and TIR modalities be denoted as \( \mathcal{M}_{\text{R}} \) and \( \mathcal{M}_{\text{T}} \), respectively. The following describes two advanced fusion strategies for combining these decisions.

\textbf{Confidence-Based Weighting with Normalization.} 
To refine the fusion process, confidence scores \( \mathcal{C}_{\text{R}} \) and \( \mathcal{C}_{\text{T}} \) reflect each modality’s reliability and serve as normalization factors. These scores are obtained through a scaling function \(\psi(\cdot)\) and normalized using \(\tau(\cdot)\):

\begin{equation}
\mathcal{C}_{\text{R}} = \tau(\psi(\mathcal{M}_{\text{R}})), \quad \mathcal{C}_{\text{T}} = \tau(\psi(\mathcal{M}_{\text{T}})).
\end{equation}

The confidence-weighted fusion result $\mathcal{Q}_{\text{fuse}}$ is:

\begin{equation}
\mathcal{Q}_{\text{fuse}} = \frac{\left(\mathcal{C}_{\text{R}} \odot \mathcal{M}_{\text{R}} + \mathcal{C}_{\text{T}} \odot \mathcal{M}_{\text{T}}\right) + \epsilon}{\mathcal{C}_{\text{R}} + \mathcal{C}_{\text{T}} + \epsilon},
\end{equation}
where \(\odot\) represents element-wise weighting, and \(\epsilon\) is a small constant to prevent division by zero, thereby stabilizing the computation.

\textbf{Hierarchical Fusion with Multi-stage Process.} 
Hierarchical fusion enhances robustness by applying both local and global fusion steps. Initially, a region-based fusion is applied independently within each modality. This local fusion step can be represented as:

\begin{equation}
\mathcal{Q}_{\text{local}} = h_{\text{local}}(\kappa_{\text{R}} \cdot \mathcal{M}_{\text{R}}, \kappa_{\text{T}} \cdot \mathcal{M}_{\text{T}}),
\end{equation}

where \( h_{\text{local}}(\cdot) \) represents the local fusion function, such as Simple Average, Confidence-Weighted Average, or Maximum Selection, and \( \kappa_{\text{R}} \) and \( \kappa_{\text{T}} \) are weighting factors specific to each modality.

After obtaining the locally fused results, a global aggregation function combines these results across regions or categories. The global fusion step is given by:

\begin{equation}
\mathcal{Q}_{\text{fuse}} = h_{\text{global}}\left( \sum_{i=1}^{N} \theta_i \, \mathcal{Q}_{\text{local}}^{(\text{i})} \right),
\end{equation}

where \( h_{\text{global}}(\cdot) \) denotes the global fusion function, \( N \) is the number of local regions or categories, and \(\theta_i\) are adaptive coefficients for each local fused region \( \mathcal{Q}_{\text{local}}^{(\text{i})} \).

This hierarchical approach provides finer control over region-specific interactions, enhancing robustness in complex scenes.

\end{enumerate}
\textbf{Experimental Observations}

We first evaluate the three fusion methods through experiments and identify feature-level fusion as the most effective approach. Building on this insight, we further optimize the combination of feature-level fusion modules to achieve the best performance.
\begin{table*}[htbp] \footnotesize
\centering
\renewcommand{\arraystretch}{0.8} 
\setlength{\tabcolsep}{6pt} 
\captionsetup{justification=centering, labelsep=period, font=bf} 
\caption{\footnotesize{Performance metrics of advanced single-modality detection models under different fusion mechanisms. The results are averaged over 100 independent runs, with the standard deviations provided. We use bold red font and underline to highlight the best results.}} 
\vspace{-0.7em}
\label{tab:fusion_comparison}
\begin{tabular}{lllc|ccccc}
\toprule
\textbf{Backbone} & \textbf{Method} & \textbf{Datasets} & & \multicolumn{5}{c}{\textbf{Fusion Strategy}} \\ 
\cmidrule(lr){5-9}
 & & & & \textbf{Pixel-Fusion} & \textbf{Feature-Fusion} & \textbf{Decision-Fusion} & \textbf{RGB-Output} & \textbf{TIR-Output} \\ 
\midrule
\multirow{8}{*}{\textbf{Resnet50}} 
 & YOLO-V5 \cite{geetha2024comparingyolov5variantsvehicle} & KAIST & & $29.31_{\pm 1.21}$ & $\mathbf{\underline{\textcolor{red}{15.17_{\pm 1.59}}}}$ & $17.37_{\pm 2.24}$ & $18.39_{\pm 1.75}$ & $17.89_{\pm 2.92}$ \\
 &         & FLIR  & & $63.53_{\pm 2.81}$ & $\mathbf{\underline{\textcolor{red}{73.22_{\pm 1.66}}}}$ & $68.71_{\pm 1.20}$ & $67.84_{\pm 1.17}$ & $68.23_{\pm 1.75}$ \\
 & CO-DETR \cite{zong2023detrscollaborativehybridassignments}& KAIST & & $26.17_{\pm 2.01}$ & $\mathbf{\underline{\textcolor{red}{14.67_{\pm 2.35}}}}$ & $16.83_{\pm 1.17}$ & $17.65_{\pm 2.55}$ & $17.17_{\pm 1.57}$ \\
 &         & FLIR  & & $62.39_{\pm 2.65}$ & $\mathbf{\underline{\textcolor{red}{78.97_{\pm 2.50}}}}$ & $69.36_{\pm 2.40}$ & $68.93_{\pm 2.12}$ & $68.35_{\pm 2.74}$ \\
 & RTMDET \cite{lyu2022rtmdetempiricalstudydesigning}  & KAIST & & $23.59_{\pm 1.64}$ & $\mathbf{\underline{\textcolor{red}{14.13_{\pm 2.58}}}}$ & $18.97_{\pm 1.15}$ & $17.36_{\pm 1.85}$ & $16.33_{\pm 1.45}$ \\
 &         & FLIR  & & $57.81_{\pm 1.97}$ & $\mathbf{\underline{\textcolor{red}{75.36_{\pm 2.31}}}}$ & $66.29_{\pm 1.71}$ & $64.32_{\pm 2.44}$ & $63.97_{\pm 2.12}$ \\
 & DINO \cite{caron2021emergingpropertiesselfsupervisedvision}    & KAIST & & $27.73_{\pm 1.98}$ & $18.93_{\pm 2.16}$ & $\mathbf{\underline{\textcolor{red}{16.67_{\pm 1.12}}}}$ & $19.57_{\pm 2.66}$ & $17.98_{\pm 1.81}$ \\
 &         & FLIR  & & $57.69_{\pm 2.87}$ & $\mathbf{\underline{\textcolor{red}{76.83_{\pm 2.33}}}}$ & $69.15_{\pm 2.14}$ & $67.96_{\pm 2.02}$ & $67.71_{\pm 2.33}$ \\
\midrule
\multirow{8}{*}{\textbf{Vit-L}} 
 & YOLO-V5 \cite{geetha2024comparingyolov5variantsvehicle} & KAIST & & $28.99_{\pm 1.98}$ & $\mathbf{\underline{\textcolor{red}{13.52_{\pm 2.18}}}}$ & $17.61_{\pm 2.32}$ & $18.02_{\pm 2.45}$ & $18.63_{\pm 2.66}$ \\
 &         & FLIR  & & $62.72_{\pm 1.91}$ & $\mathbf{\underline{\textcolor{red}{72.67_{\pm 2.71}}}}$ & $68.72_{\pm 1.99}$ & $68.12_{\pm 2.47}$ & $68.61_{\pm 2.99}$ \\
 & CO-DETR \cite{zong2023detrscollaborativehybridassignments} & KAIST & & $27.95_{\pm 2.53}$ & $\mathbf{\underline{\textcolor{red}{13.63_{\pm 1.53}}}}$ & $16.85_{\pm 2.34}$ & $18.36_{\pm 1.90}$ & $17.17_{\pm 2.97}$ \\
 &         & FLIR  & & $63.30_{\pm 2.89}$  & $\mathbf{\underline{\textcolor{red}{76.55_{\pm 2.35}}}}$ & $70.72_{\pm 1.52}$ & $67.27_{\pm 2.95}$ & $69.65_{\pm 2.11}$ \\
 & RTMDET \cite{lyu2022rtmdetempiricalstudydesigning}  & KAIST & & $22.60_{\pm 2.32}$  & $15.11_{\pm 2.65}$ & $\mathbf{\underline{\textcolor{red}{14.53_{\pm 2.95}}}}$ & $16.34_{\pm 2.15}$ & $16.19_{\pm 2.98}$ \\
 &         & FLIR  & & $56.75_{\pm 2.78}$ & $\mathbf{\underline{\textcolor{red}{74.39_{\pm 2.19}}}}$ & $66.79_{\pm 1.61}$ & $65.63_{\pm 2.22}$ & $65.07_{\pm 1.68}$ \\
 & DINO \cite{caron2021emergingpropertiesselfsupervisedvision}    & KAIST & & $26.97_{\pm 2.68}$ & $\mathbf{\underline{\textcolor{red}{12.21_{\pm 2.95}}}}$ & $15.54_{\pm 1.95}$ & $19.89_{\pm 1.85}$ & $18.08_{\pm 2.34}$ \\
 &         & FLIR  & & $56.11_{\pm 1.89}$ & $\mathbf{\underline{\textcolor{red}{77.12_{\pm 1.99}}}}$ & $70.61_{\pm 1.62}$ & $68.37_{\pm 2.95}$ & $69.97_{\pm 2.11}$ \\
\bottomrule
\end{tabular}
\end{table*}

\begin{enumerate}
\item[\ding{72}] \textbf{Fusion Method Experiments.}
In our preliminary experiments, we compared the effects of the three feature fusion methods on the improved multispectral model. The experimental results can be found in Table \ref{tab:fusion_comparison}. It is evident that using different fusion methods had a significant impact on the detection accuracy of the optimized model.
\begin{enumerate}[label=(\roman*):,leftmargin=12pt]
    \item \textbf{Observations on Pixel-Level Fusion.} 
    Pixel-level fusion exhibits lower stability and detection accuracy compared to single-modality detection on most datasets, with only slight improvements observed in a few specific cases. This may be attributed to the fact that pixel-level fusion combines the dual-light images at the input stage, introducing a significant amount of redundant information and noise. As a result, the model struggles to effectively learn the key features from each modality.
    
    \item \textbf{Observations on Feature-Level Fusion.} 
    Compared to single-modality detection, feature-level fusion demonstrated significant improvements in both stability and detection accuracy across most datasets. This is likely due to the fact that feature-level fusion effectively utilizes high-level features extracted by the backbone, allowing for efficient fusion while minimizing redundant features and preserving as much valuable information as possible.
    
    \item \textbf{Observations on Decision-Level Fusion.} 
    Compared to single-modality detection, decision-level fusion can improve accuracy to some extent, but it demonstrates instability with certain methods, such as the RTMDet framework \cite{lyu2022rtmdetempiricalstudydesigning}. This instability may stem from the fact that decision-level fusion processes RGB and TIR modality information independently, merging them only at the decision stage. Consequently, this approach struggles to effectively leverage complementary information between the two modalities, especially in scenarios where such information is crucial, like varying weather conditions or significant changes in viewpoints.
\end{enumerate}

\item[\ding{72}] \textbf{Feature-Fusion Experiments.}
To determine the most effective fusion strategy, we selected the best-performing feature-level fusion method from prior experiments for further analysis. Using single-modality detection models as baselines, we introduced the NIN and ICFE modules under different input modalities. This approach enabled a systematic evaluation of their contributions to feature representation and fusion performance. Key results are shown in Figure \ref{fig:icfe}, along with notable findings.

\begin{figure*}[h]
    \centering
        \begin{minipage}{\linewidth}
            \centering
            \begin{minipage}{0.5\linewidth}
                \centering
                \includegraphics[width=\linewidth]{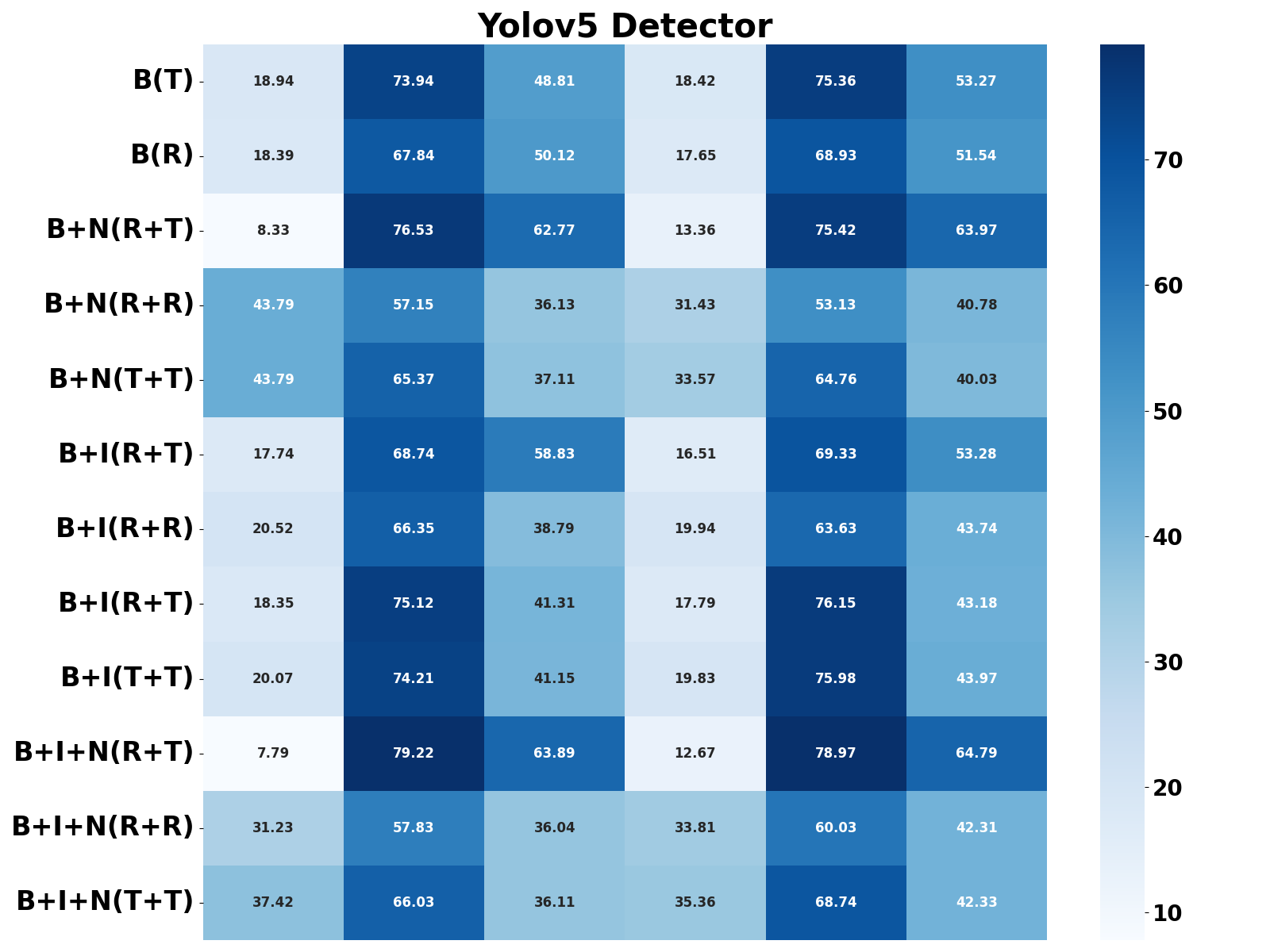} 
            \end{minipage}%
            \hfill
            \begin{minipage}{0.5\linewidth}
                \centering
                \includegraphics[width=\linewidth]{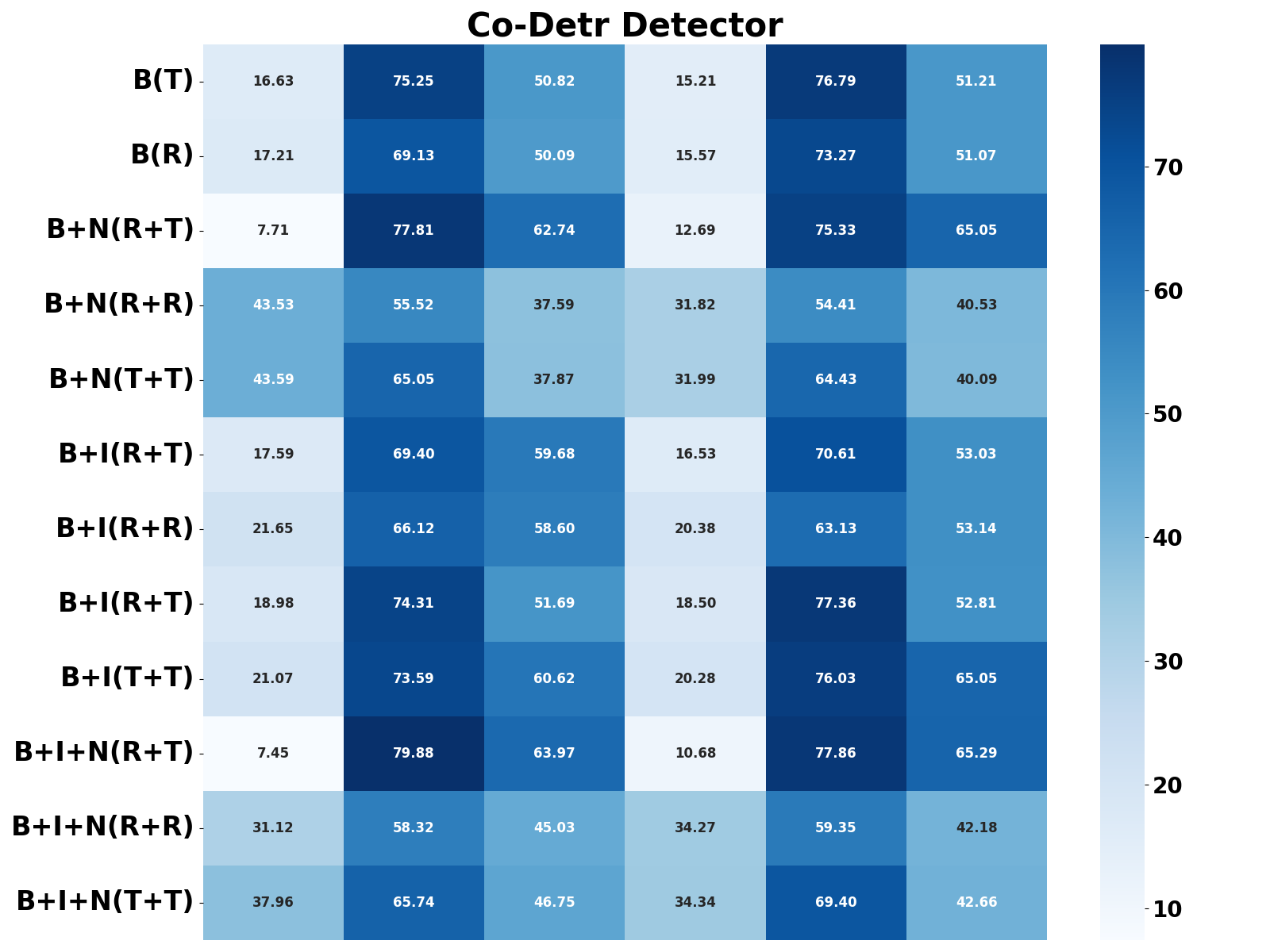} 
            \end{minipage}
        \end{minipage}
    \caption{Performance metrics obtained from 100 independent repetitions on the KAIST, FLIR, and DroneVehicle datasets using different backbones and feature fusion modules. The letter B represents the baseline, I represents the ICFE module, and N represents the NIN module, while the content in parentheses indicates the modality input to the fusion module. The left images show the results from the experiments using the Yolov5 detector, while the right images present the results from the experiments using the Co-Detr detector. Each column in the figures, from left to right, represents: the results with Resnet50 as the backbone on the KAIST, FLIR, and DroneVehicle datasets, followed by the results with Vit-L as the backbone on the same datasets.}
    \label{fig:icfe}
\end{figure*}

\begin{enumerate}[label=(\roman*):,leftmargin=12pt]

    \item \textbf{Observations on Datasets.} After applying fusion modules, all detection frameworks showed varying degrees of improvements. Notably, on datasets with significant changes in lighting conditions, shadows, and viewpoints (e.g., the FLIR dataset), both the NIN-structured fusion module and the ICFE-structured fusion module exhibited more pronounced performance. This enhancement is likely attributable to the fact that in scenarios where there are substantial differences between the two modalities, complementary information plays a crucial role in improving detection accuracy, which highlights the effectiveness of the fusion modules.
    
    \item \textbf{Observations on Fusion Modules.} We found that different fusion module architectures exhibit high sensitivity to various backbone networks. Specifically, in detection networks using Resnet50 as the backbone, the NIN-structured fusion module showed notable improvements in detection accuracy. On the other hand, for backbones based on the Vit-L structure, the ICFE module demonstrated better performance when fusing data from the RGB and TIR channels. This difference in performance may be attributed to the fact that Resnet50 is a convolution-based architecture, where the NIN module effectively fuses local features, maintaining the continuity and consistency of convolutional features, thus leading to better results. In contrast, Vit-L excels at capturing global features, and the ICFE module, with its cross-feature and attention mechanisms, further enhances the fusion of global information, resulting in superior performance.
    
    \item \textbf{Observations on the ICFE Fusion Module Branches.} 
    For the branch inputs of the ICFE module, we experimented with various connection methods, as illustrated in Figure \ref{fig:icfe}. The experimental results show that using the ICFE module alone for fusion, regardless of the connection method, failed to consistently improve the detection accuracy. This outcome may be attributed to the fact that when only a single module is used for fusion with inputs from the same modality, the ICFE module may repeatedly amplify background noise or irrelevant features, causing the model to focus excessively on the noise rather than the target, thereby reducing detection performance. Furthermore, when inputs from different modalities (RGB and TIR) are used, their features are not deeply fused or integrated (e.g., through NIN’s nonlinear transformation), meaning the complementary information between modalities is not fully leveraged.
    
    We further attempted to add an NIN connection structure after the iterative ICFE module, using different input methods. The experimental results indicate that using the R+T+NIN connection significantly improves the detection accuracy, while the R+R and T+T configurations, following NIN extraction, resulted in poorer performance. This is likely due to that the NIN module can more finely integrate and fuse cross-modality features, leading to notable improvements in detection performance.
    
    \item \textbf{Observations on Robustness.} The experimental results indicate that different input configurations (e.g., R+T, R+R, T+T) have a significant impact on the model's robustness. When using the same modality inputs (R+R or T+T), the model's detection performance tends to be unstable and more susceptible to background noise. In contrast, when using the R+T combination, especially when coupled with the NIN module for feature fusion, the model demonstrates significantly higher robustness across various environmental conditions. These findings suggest that the complementary information between modalities plays a crucial role in enhancing the model’s ability to withstand environmental uncertainty and noise interference.
\end{enumerate}
\end{enumerate}

\subsection{Dual-Modality Data Augmentation}
\textbf{Formulations.} Dual-modality data augmentation is a vital technique for enhancing the performance of multispectral object detection models. By applying consistent or complementary transformations to both modalities during training, this approach not only ensures the correlation between features from the two data sources but also enables the simulation of specific test scenarios (e.g., low-light conditions or small samples). Additionally, it effectively addresses information loss caused by feature dimensionality reduction, particularly in cases where the data distributions of the two modalities differ significantly. Mainstream dual-modality data augmentation strategies can be broadly categorized into three types: \textbf{Geometric Transformations, Pixel-Level Transformations, and Multimodal-Specific Enhancements}. These strategies will be detailed in the following sections.

\begin{enumerate}[leftmargin=*]
\item[\ding{115}] \textbf{Geometric Transformations.} 
Geometric transformation strategies involve a range of spatial modifications designed to maximize the geometric diversity of training samples, enabling the model to generalize more effectively to varied object poses, orientations, scales, and viewpoints. The overall approach to geometric transformation strategies is outlined below, with most transformations formulated based on the following equation. Let the input image be represented by $\mathcal{I}$, the processed image by $\mathcal{I}'$, and the geometric transformation function by $\mathcal{F}_g$. This transformation can be formalized as:

\begin{equation}
\mathcal{I}' = \mathcal{F}_g(\mathcal{I}) = \rho \cdot \mathcal{I} + \Upsilon,
\end{equation}

where $\rho$ denotes the composite affine transformation matrix, which integrates non-uniform scaling, complex rotation, and controlled mirroring. The $\Upsilon$ represents the non-linear offset coefficient.

The matrix $\rho$ can be decomposed as:

\begin{equation}
\rho = \mathcal{S}(c_x, c_y) \cdot \mathcal{R}(\theta) \cdot \mathcal{U}_\ell(\phi) \cdot \mathcal{E}(t_x, t_y),
\end{equation}

where each component transformation is defined as follows:

- $\mathcal{S}(c_x, c_y)$ represents a non-uniform scaling matrix, applying differential scaling along the $x$ and $y$ axes:
\begin{equation}
\mathcal{S}(c_x, c_y) = \begin{bmatrix} c_x & 0 \\ 0 & c_y \end{bmatrix},
\end{equation}

where $c_x$ and $c_y$ are the horizontal and vertical scaling factors, respectively, which may vary based on context-specific augmentation parameters.

- $\mathcal{R}(\theta)$ denotes the rotation matrix, which rotates the image by an angle $\theta$ in the 2D plane:
\begin{equation}
\mathcal{R}(\theta) = \begin{bmatrix} \cos(\theta) & -\sin(\theta) \\ \sin(\theta) & \cos(\theta) \end{bmatrix}.
\end{equation}

- $\mathcal{U}_\ell(\phi)$ represents the mirroring transformation, capable of inducing horizontal or vertical flips, denoted as follows:

{\small
\begin{equation}
\mathcal{U}_\ell(\phi) = \left\{ 
\begin{array}{ll}
\begin{bmatrix} -\cos(\phi) & 0 \\ 0 & \cos(\phi) \end{bmatrix}, & \text{if } \ell = \text{horizontal} \\[12pt]
\begin{bmatrix} \cos(\phi) & 0 \\ 0 & -\cos(\phi) \end{bmatrix}, & \text{if } \ell = \text{vertical}
\end{array},
\right.
\end{equation}
}

where $\phi$ is a stochastic parameter controlling the mirroring type, potentially following a probabilistic distribution to introduce randomness into the flipping process. This matrix may be further generalized to incorporate combinations of horizontal and vertical mirroring transformations, represented as:

{\small
\begin{equation}
\mathcal{U}(\phi_h, \phi_v) = \begin{bmatrix} \cos(\phi_h) \cdot \cos(\phi_v) & 0 \\ 0 & \cos(\phi_h) \cdot \cos(\phi_v) \end{bmatrix}.
\end{equation}
}

- $\mathcal{E}(t_x, t_y)$ is the translation matrix, introducing positional shifts along the $x$ and $y$ axes:
\begin{equation}
\mathcal{E}(t_x, t_y) = \begin{bmatrix} 1 & 0 & t_x \\ 0 & 1 & t_y \\ 0 & 0 & 1 \end{bmatrix},
\end{equation}

where $t_x$ and $t_y$ represent horizontal and vertical translations, respectively. These shifts may vary based on contextual constraints to simulate different spatial orientations.

\item[\ding{115}] \textbf{Pixel-Level Transformations.}  
Pixel-level transformation strategies modify the pixel values of an image, such as by adding noise, adjusting colors, or altering contrast, to simulate various imaging conditions. This enhances the model's robustness to lighting variations, noise, and diverse environmental factors. The following introduces pixel-level transformation strategies, with most transformations adhering to the approach outlined below. Let the pixel matrix of the image be $\mathcal{P}$, the transformation can be expressed through the following steps:

\textbf{Noise Addition.} 
To simulate sensor noise or environmental interference, Gaussian noise $N(\sigma)$ with a standard deviation of $\sigma$ is added to the pixel matrix:
\begin{equation}
\mathcal{P}_{\text{noise}} = \mathcal{P} + \mathcal{N}(\sigma),
\end{equation}
where $\mathcal{N}(\sigma)$ represents Gaussian noise with variance $\sigma^2$.

\textbf{Color Adjustment.} 
To simulate different lighting conditions or sensor biases, color adjustment is applied using a scaling factor $\alpha$:
\begin{equation}
\mathcal{P}_{\text{color}} = \mathcal{C}(\alpha) \cdot \mathcal{P}_{\text{noise}},
\end{equation}
where $\alpha$ is the color adjustment factor that controls the brightness or saturation of each channel.

\textbf{Contrast Adjustment.} 
To enhance or reduce image details, contrast adjustment is applied using a contrast factor $\beta$:
\begin{equation}
\mathcal{P}_{\text{contrast}} = \mathcal{D}(\beta) \cdot (\mathcal{P}_{\text{color}} - \mu) + \mu,
\end{equation}
where $\beta$ is the contrast adjustment factor and $\mu$ is the mean pixel value used for centering the pixel matrix.

\textbf{Final Pixel Transformation.} 
The final pixel transformation combines all the above operations:
\begin{equation}
\mathcal{P}' = \mathcal{D}(\beta) \cdot \mathcal{C}(\alpha) \cdot (\mathcal{P} + \mathcal{N}(\sigma)).
\end{equation}

\item[\ding{115}]\textbf{Multimodal-Specific Enhancements.} 
This class of strategies focuses on the unique characteristics of dual-light data, employing dual-channel synchronized or complementary enhancements tailored to specific test scenarios. By applying different augmentation methods to each modality, these strategies effectively enhance the cooperative performance of multimodal images and improve accuracy in targeted detection scenarios. Let the RGB image be denoted as \( \mathcal{I}_{\text{R}} \) and the TIR image as \( \mathcal{I}_{\text{T}} \). The multimodal-specific enhancement can be expressed as:

\begin{equation}
\begin{bmatrix}
\mathcal{I}'_{\text{R}} \\
\mathcal{I}'_{\text{T}}
\end{bmatrix} = \tau \left( \mathcal{I}_{\text{R}}, \mathcal{I}_{\text{T}} \right),
\end{equation}

where \( \tau \) represents the multimodal enhancement function, which may include cross-modal alignment and modality-specific feature enhancement. The \( \mathcal{I}'_{\text{R}} \) and \( \mathcal{I}'_{\text{T}} \) represent the enhanced RGB and TIR images, respectively. Specifically, the enhancement process can be further detailed as:

\begin{equation}
\begin{bmatrix}
\mathcal{I}'_{\text{R}} \\
\mathcal{I}'_{\text{T}}
\end{bmatrix} = 
\begin{bmatrix}
\varpi_{\text{R}} \left( \mathcal{I}_{\text{R}}, \mathcal{A}_{\text{T}} \cdot \mathcal{L}(\mathcal{I}_{\text{T}}) \right) \\
\varpi_{\text{T}} \left( \mathcal{I}_{\text{T}}, \mathcal{A}_{\text{R}} \cdot \mathcal{L}(\mathcal{I}_{\text{R}}) \right)
\end{bmatrix}.
\end{equation}

The functions \( \varpi_{\text{R}} \) and \( \varpi_{\text{T}} \) denote modality-specific enhancement operations applied to the input images, incorporating their corresponding aligned features. The matrices \( \mathcal{A}_{\text{T}} \) and \( \mathcal{A}_{\text{R}} \) are modality-specific alignment matrices, while \( \mathcal{L}(\mathcal{I}_{\text{modality}}) \) serves as the feature extraction function that identifies crucial features within each image for optimized information integration.

\end{enumerate}

\noindent\textbf{Experimental observations} 

\begin{figure*}[h]
    \centering
        \begin{minipage}{\linewidth}
            \centering
            \begin{minipage}{0.49\linewidth}
                \centering
                \includegraphics[width=\linewidth]{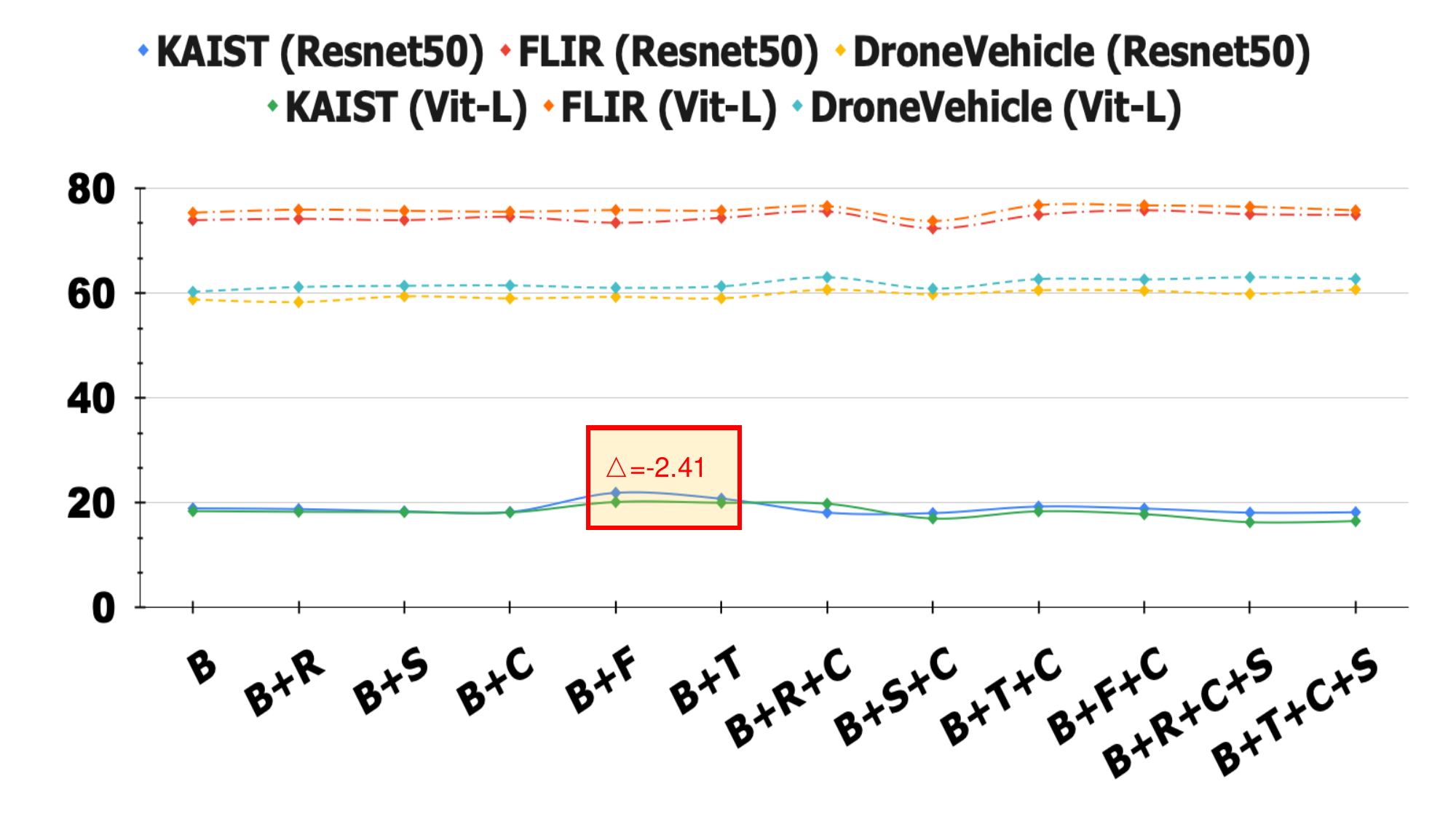} 
            \end{minipage}%
            \hfill
            \begin{minipage}{0.49\linewidth}
                \centering
                \includegraphics[width=\linewidth]{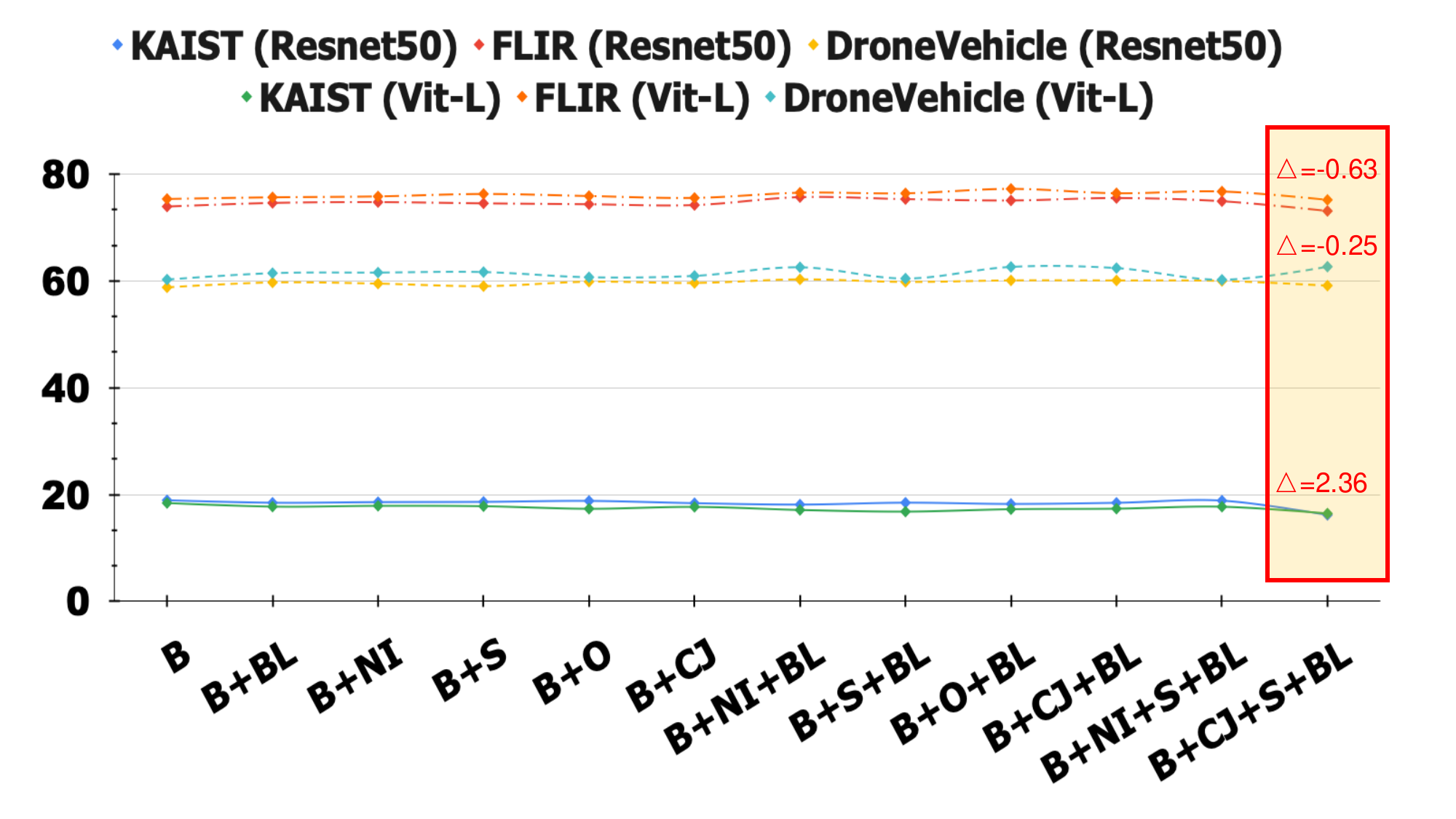} 
            \end{minipage}
        \end{minipage}
        \vspace{-1.2em}
    \caption{The performance of general geometric and pixel-level augmentations (using different backbones) on the KAIST, FLIR, and DroneVehicle datasets. The left figure illustrates the results of various geometric augmentations, where B denotes the baseline, R represents random rotation, S signifies multi-scale scaling, C stands for random cropping, F corresponds to random flipping, and T indicates random translation. The right figure presents the results of general pixel-level augmentations, with B as the baseline, BL for random blurring, NI for noise injection, S for random sharpening, O for random occlusion, and CJ for color jittering. $\triangle$ represents the mean performance difference between this method and the baseline.} 
    \vspace{-1.2em}
    \label{fig:g1}
\end{figure*}

\begin{figure}[htbp]
    \centering
    \includegraphics[width=1\linewidth]{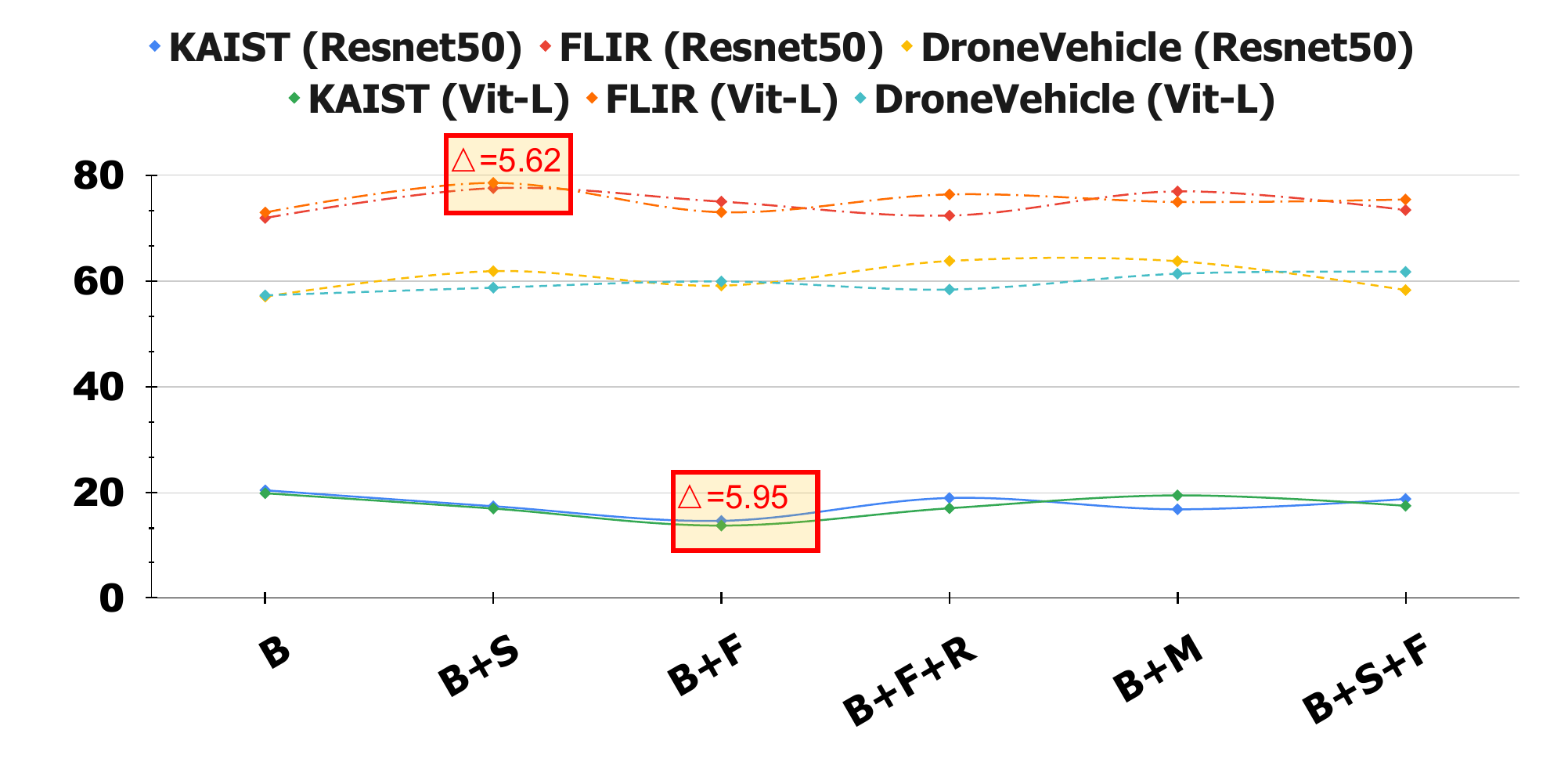} 
    \caption{The performance metrics of different augmentation strategies on small sample sets. $\triangle$ represents the mean performance difference between this method and the baseline.
    In this figure,  B represents the baseline, S denotes Stitcher \cite{chen2021dynamicscaletrainingobject}, F stands for Fastmosaic \cite{kumar2020yolov3}, R represents Region Resampling, and M indicates Small-Object Magnification.}
    \label{fig:sp2} 
\end{figure}

Based on the single-modality object detection model Co-Detr, we made adaptive modifications to construct a baseline model suitable for multispectral object detection. As multispectral object detection augmentation strategies often need to adapt to specific application scenarios, test set sample characteristics, and varying weather and lighting conditions, we first conducted experiments exploring a set of synchronized augmentation techniques focused on geometric and pixel-level transformations. The experimental results are shown in Figure \ref{fig:g1}. Building upon these methods, we further investigate specific augmentation strategies tailored to the unique characteristics of dual-light samples. The experimental results are shown in Figures \ref{fig:sp1} and \ref{fig:sp2}.


\begin{enumerate}
\item[\ding{72}] 
\textbf{General Augmentation Strategy Experiments.} In this section, we conducted dual-channel synchronized augmentation experiments using various geometric and pixel-level strategies, revealing several key insights.

\begin{enumerate}[label=(\roman*):,leftmargin=12pt]

    \item \textbf{Observations on Geometric Transformations.} 
    The experimental data indicates that applying a combination of random rotation, multi-scale scaling, and random cropping results in performance improvements across multiple datasets. However, strategies such as random flipping and random translation show poorer performance on the KAIST dataset. This could be attributed to the fact that the combination of random rotation, multi-scale scaling, and random cropping effectively simulates samples from various perspectives and angles, thus enhancing the model's ability to adapt to different viewpoints, angles, scales, and deformations. On the other hand, strategies like flipping and translation may produce illogical images for certain samples (e.g., flipping upright pedestrians in the KAIST dataset leads to unnatural postures), which disrupts the inherent distribution patterns and modality alignment in some datasets, negatively affecting detection performance.
    \item \textbf{Observations on Pixel-Level Transformations.} 
    The overall performance improvements from pixel-level augmentation strategies are less significant compared to geometric transformations or spatial alignment methods. For instance, even the most effective combination in our experiments yielded only a 2.5\% increase in recognition accuracy over the baseline, which is relatively modest when compared to methods such as feature fusion. Besides, a large number of pixel-level augmentation strategies (three or more) exhibit high sensitivity to different datasets. Specifically, we observed that the combination of Color Jitter+Random Sharpening+Random Blurring significantly improved recognition accuracy on the KAIST dataset, but the same combination performed poorly on the FLIR dataset. When more than four pixel-level augmentation strategies were applied, recognition accuracy often plateaued or even decreased across multiple datasets.
    \end{enumerate}
    \end{enumerate}

\begin{enumerate}
\item[\ding{72}] \textbf{Experiments on Unique Augmentation Strategies.} 
For specific scenarios, such as low-light/nighttime conditions and very small sample cases, we selected 500 images from the original dataset that exhibit these characteristics for targeted testing. We experimented with various combinations of dual-channel augmentation strategies, which includes dual-channel synchronized augmentation and complementary augmentation. Below are some interesting observations:
\begin{figure}[h]
    \centering
    \begin{minipage}{\linewidth}
        \centering
        \includegraphics[width=1\linewidth]{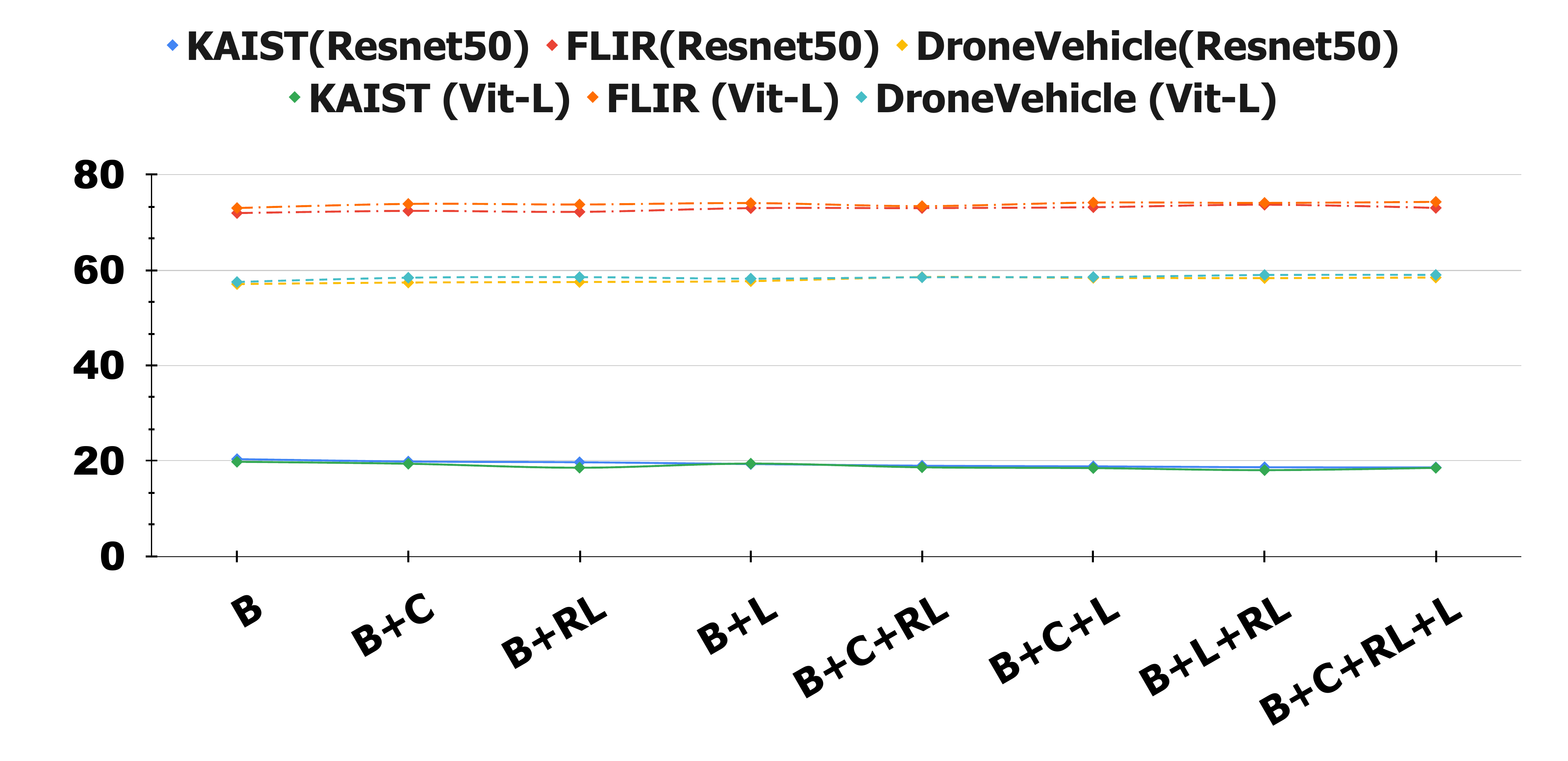}
        \label{fig:enter-label-1}
    \end{minipage}
    \vfill
    \begin{minipage}{\linewidth}
        \centering
        \includegraphics[width=1\linewidth]{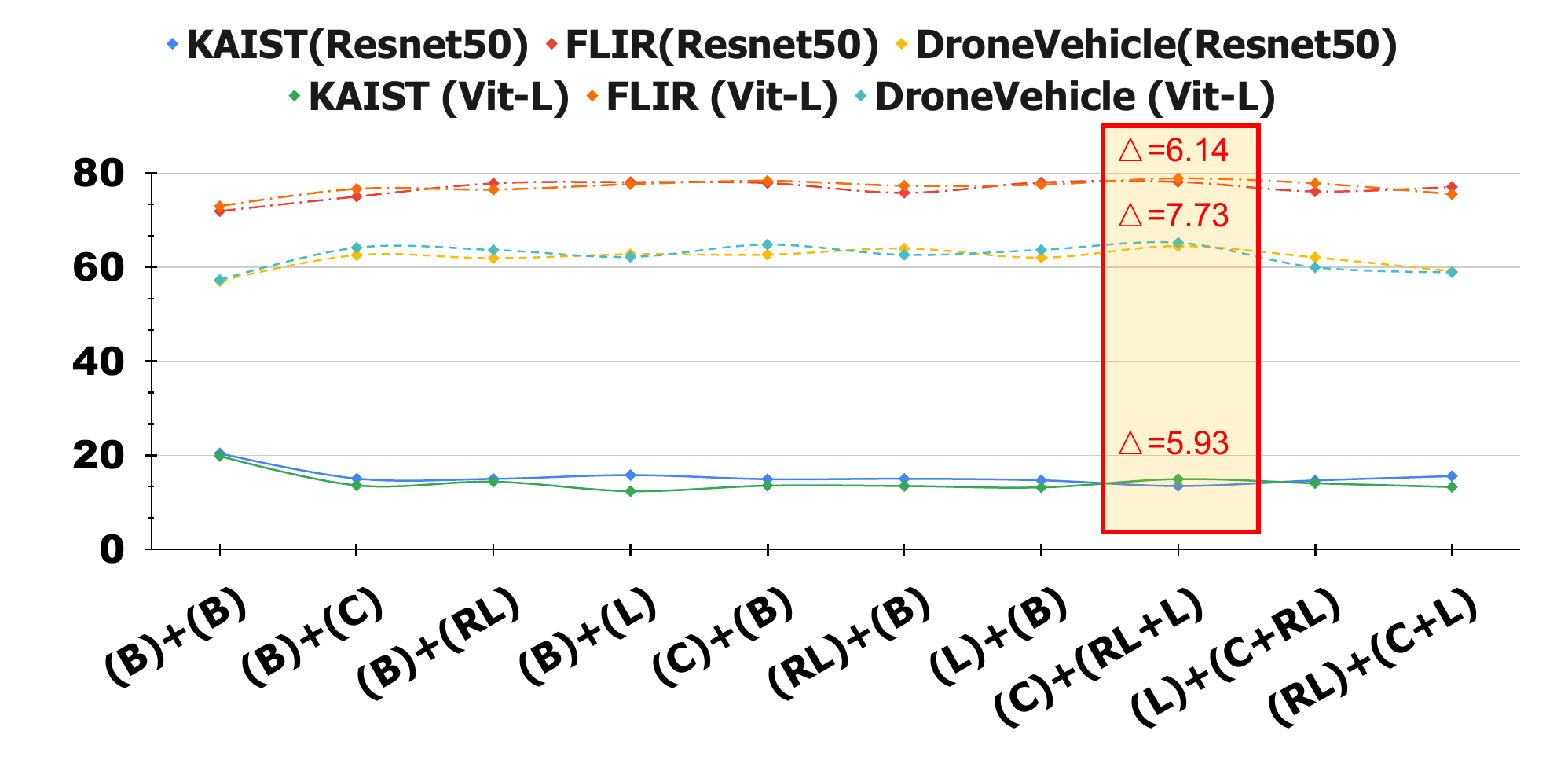}
        \label{fig:enter-label-2}
    \end{minipage}
    \vspace{-1.2em}
    \caption{The performance metrics of different augmentation strategies applied to nighttime/low-light samples. The top image shows the results using dual-channel synchronized augmentation, while the bottom image displays results with dual-channel complementary augmentation. In both images, B represents the baseline, C stands for CLAHE, RL denotes random lighting, and L indicates light enhancement. In the bottom image, each set of parentheses indicates the specific augmentation strategies applied to each modality, with the order representing the RGB and TIR channels, respectively. $\triangle$ represents the mean performance difference. }
    \label{fig:sp1}
    \vspace{-1.2em}
\end{figure}

\begin{enumerate}[label=(\roman*):,leftmargin=12pt]
    \item \textbf{Observations on Strategies for Nighttime/Low-Light Samples.}
    We conducted experiments comparing both synchronous and complementary augmentation strategies to identify the most effective combination for enhancing performance in low-light conditions. We found that complementary augmentation outperforms synchronous augmentation in improving overall recognition accuracy. This improvement is particularly pronounced in low-light conditions, where the strengths of complementary augmentation are more evident. Specifically, in low-light environments, the RGB modality tends to suffer from information loss, such as reduced contrast and increased noise, while the TIR modality, which captures thermal radiation, continues to provide stable target information even in the absence of illumination. Thus, adopting a complementary augmentation strategy allows each modality to better leverage its respective strengths. Besides, The complementary augmentation combination of random lighting and light enhancement for the TIR channel, paired with CLAHE for the RGB channel, achieved excellent results across all datasets. This success can be attributed to the complementary strategy’s ability to enhance the adaptability of the RGB channel to varying lighting conditions, while simultaneously improving the clarity of edges and shapes in the infrared samples.

    \item \textbf{Observations on Strategies for Small Samples.}
    From the experimental data, it is evident that the augmentation strategy improves recognition accuracy. Specifically, the stitching operation proved to be highly effective in addressing the problem of very small samples individually, while the other two augmentation techniques did not consistently improve recognition accuracy. Independent use of both Stitcher and Fastmosaic led to notable improvements in recognition accuracy. In particular, Fastmosaic was the preferred choice for large-scale datasets (such as KAIST), while Stitcher performed better on more complex datasets (such as FLIR). Interestingly, when the two methods were combined, recognition accuracy decreased compared to their individual use. This outcome could be attributed to an imbalance in data distribution caused by the combination, which failed to provide the model with additional useful information.
\end{enumerate}
\end{enumerate}

\subsection{Registration Alignment} 
\textbf{Formulations.} 
In multispectral object detection tasks, factors such as sensor viewpoints, resolution discrepancies, and varying weather conditions can lead to spatial misalignment between RGB and TIR images. Such misalignment often introduces inconsistencies during feature fusion, thereby degrading detection performance. To address these issues, researchers have developed various registration and alignment strategies, which can be broadly categorized into \textbf{Feature Alignment-Based Methods and Feature Fusion-Based Methods}. By applying these registration techniques at different stages of training and testing, the alignment between RGB and TIR images can be effectively improved, significantly enhancing recognition accuracy. The following sections provide a detailed discussion of these two categories.

\begin{enumerate}[leftmargin=*]
\item[\ding{115}] \textbf{Feature Alignment-Based Methods.}  
The main goal of these methods is to address spatial misalignment between RGB and TIR images through precise feature matching and alignment. The Loftr approach exemplifies this objective by leveraging a Transformer-based architecture to achieve pixel-level feature matching between RGB and TIR images, allowing for high-precision geometric alignment \cite{sun2021loftrdetectorfreelocalfeature}. This approach enables the calculation of transformation parameters (such as affine or perspective transformations) that can be applied to register the images effectively.

Let the RGB image be denoted as \( \mathcal{I}_{\text{R}} \) and the TIR image as \( \mathcal{I}_{\text{T}} \). The coarse and fine features extracted from these images are represented as \( \Phi_{\text{R}} \) and \( \Phi_{\text{T}} \), respectively. The matching function \( \varrho_{\text{m}}(\Phi_{\text{R}}, \Phi_{\text{T}}) \) can be formulated as follows:

{\small
\begin{equation}
\varrho_{\text{m}}(\Phi_{\text{R}}, \Phi_{\text{T}}) = 
\left\{ (p_i, q_j) \mid \hat{p}_i = \sigma \left( \frac{\Phi_{\text{R}}(p_i) \cdot \Phi_{\text{T}}(q_j)}{\tau} \right) \right\},
\end{equation}
}

where \( (p_i, q_j) \) represents matched point pairs across RGB and TIR modalities, and \( \tau \) is a temperature parameter controlling the similarity distribution. \( \sigma \) denotes the softmax function, and \( \hat{p}_i \) represents the point with the highest matching score in the RGB modality for each \( q_j \) in the TIR modality.

The geometric transformation \( \mathcal{T}_g \) is then estimated based on these matched points by minimizing a distance-based objective:

\begin{equation}
\theta^* = \arg\min_{\theta} \sum_{(p_i, q_j) \in M} \left\| \mathcal{T}_g(p_i,\theta) - q_j \right\|^2,
\end{equation}
where \( \mathcal{T}_g(p_i,\theta) \) represents the transformed location of \( p_i \) in the TIR image space, with \( \theta \) containing transformation parameters for an affine or homography matrix \(\mathcal{A} \) and translation vector \( \mathcal{B} \). Once optimized, the transformation can be applied to obtain the aligned image:

\begin{equation}
\mathcal{I}_{\text{aligned}}(x,y) = \mathcal{T}_g(I_{\text{T}}, \theta^*) = \mathcal{A}(\theta^*) \cdot \mathcal{I}_{\text{T}} + \mathcal{B}.
\end{equation}

To further improve the registration accuracy, a joint loss that combines a feature consistency loss and an alignment loss are introduced, expressed as:

{\small
\begin{equation}
\mathcal{L} = \lambda_1\sum_{(p_i, q_j) \in M} \left\| \Phi_{\text{R}}(p_i) - \Phi_{\text{T}}(q_j) \right\|^2 + \lambda_2 \mathcal{L}_{\text{alignment}}(\theta),
\end{equation}
}

where \( \mathcal{L}_{\text{alignment}}(\theta) \) measures the alignment quality based on transformation parameters, and \( \lambda_1 \) and \( \lambda_2 \) are weighting coefficients to balance the two loss terms. This method demonstrates exceptional alignment capabilities in scenes with pronounced parallax and varying viewpoints, enabling efficient image registration.

\item[\ding{115}] \textbf{Feature Fusion-Based Methods.} 
Feature fusion-based methods aim to effectively combine deep RGB and TIR features to generate a fused image, thereby achieving modality alignment. SuperFusion is a prime example, employing a multilevel fusion strategy that includes data-level transformation, feature-level attention mechanisms, and final Bird’s Eye View (BEV) alignment \cite{dong2024superfusionmultilevellidarcamerafusion}.

Given an RGB image \( \mathcal{I}_{\text{R}} \) and a TIR image \( \mathcal{I}_{\text{T}} \), the process begins by extracting feature maps \( \mathcal{X}_{\text{R}} \) and \( \mathcal{X}_{\text{T}} \) through separate convolutional backbones. To enhance depth perception, a sparse depth map \( \mathcal{D}_{\text{sparse}} \) is generated by projecting TIR depth information into the RGB image plane. A completion function \( \mathcal{S}(\cdot) \) then generates a dense depth map \( \mathcal{D}_{\text{dense}} \):
\begin{equation}
\mathcal{D}_{\text{dense}} = \mathcal{S}(\mathcal{D}_{\text{sparse}}).
\end{equation}

In the feature fusion stage, cross-attention is used to align features from both modalities, where RGB features \( \mathcal{X}_{\text{R}} \) guide the enhancement of TIR features \( \mathcal{X}_{\text{T}} \). The cross-attention matrix \( \mathcal{H} \) incorporates depth information from \( \mathcal{D}_{\text{dense}} \) and is defined as:

{\small
\begin{equation}
\mathcal{H} = \sigma \left( \frac{\mathcal{Q} \mathcal{K}^T \cdot \mathcal{D}_{\text{dense}}}{\sqrt{d}} \right), \quad \mathcal{Q} = \mathcal{W}_\text{q} \mathcal{X}_{\text{R}}, \quad \mathcal{K} = \mathcal{W}_\text{k} \mathcal{X}_{\text{T}},
\end{equation}
}

where \( \mathcal{W}_\text{q} \) and \( \mathcal{W}_\text{k} \) are learned weights and \( d \) is a scaling factor, and \( \sigma \) denotes the softmax function. This mechanism aligns features across modalities by using depth information to refine attention, allowing RGB features to enrich TIR information in the fused representation.

The resulting attention matrix \( \mathcal{H} \) is then used to enhance the TIR features:
\begin{equation}
\mathcal{X}_{\text{T}}' = \mathcal{H} \cdot \mathcal{V}, \quad \mathcal{V} = \mathcal{W}_\text{v} \mathcal{X}_{\text{T}},
\end{equation}

where \( \mathcal{W}_\text{v} \) is a learned weight matrix for generating the value matrix \( \mathcal{V} \), and \( \mathcal{X}_{\text{T}}' \) represents the TIR features enhanced by the RGB guidance.

Finally, a BEV alignment module refines the fused feature map by learning a flow field \( \Delta \) to warp RGB features \( \mathcal{X}_{\text{R}} \), achieving better alignment with the enhanced TIR features \( \mathcal{X}_{\text{T}}' \). The aligned RGB image \( \mathcal{I}_{\text{aligned}} \) can be expressed as:

\begin{equation}
\mathcal{I}_{\text{aligned}}(x, y) = \sum_{x', y'} \mathcal{X}_{\text{T}}'(x', y') \, w(x, y, x', y', \Delta),
\end{equation}

where \( w(x, y, x', y', \Delta) \) represents the bilinear interpolation weights based on the flow field \( \Delta \) to adjust the alignment features. The interpolation weights \( w \) can be defined as:

\begin{equation}
w(x, y, x', y', \Delta) = \prod_{i \in \{x, y\}} \max \big(0, 1 - |i' - i - \Delta_i|\big).
\end{equation}

These weights ensure that the spatial position of the RGB features is precisely adjusted according to the flow field \( \Delta \), allowing for better alignment with the TIR features.

The entire process is optimized by a joint loss function \( \mathcal{L} \) that combines feature consistency and alignment error terms, weighted by \( \lambda_1 \) and \( \lambda_2 \):

\begin{equation}
\mathcal{L} = \lambda_1 \, \mathcal{L}_{\text{feature}} + \lambda_2 \, \mathcal{L}_{\text{alignment}},
\end{equation}

where the feature consistency term \( \mathcal{L}_{\text{feature}} = \sum_{(p_i, q_j) \in \text{M}} \left\| \mathcal{X}_{\text{R}}(p_i) - \mathcal{X}_{\text{T}}'(q_j) \right\|^2 \) minimizes the difference between matched feature pairs \( (p_i, q_j) \) in set \( \text{M} \), and the alignment term \( \mathcal{L}_{\text{alignment}} = \sum_{x, y} \left\| \Delta(x, y) - \Delta^*(x, y) \right\|^2 \) measures the deviation from the ideal alignment \( \Delta^* \).
\end{enumerate}

\begin{figure*}[htbp]
    \centering
    \includegraphics[width=\textwidth]{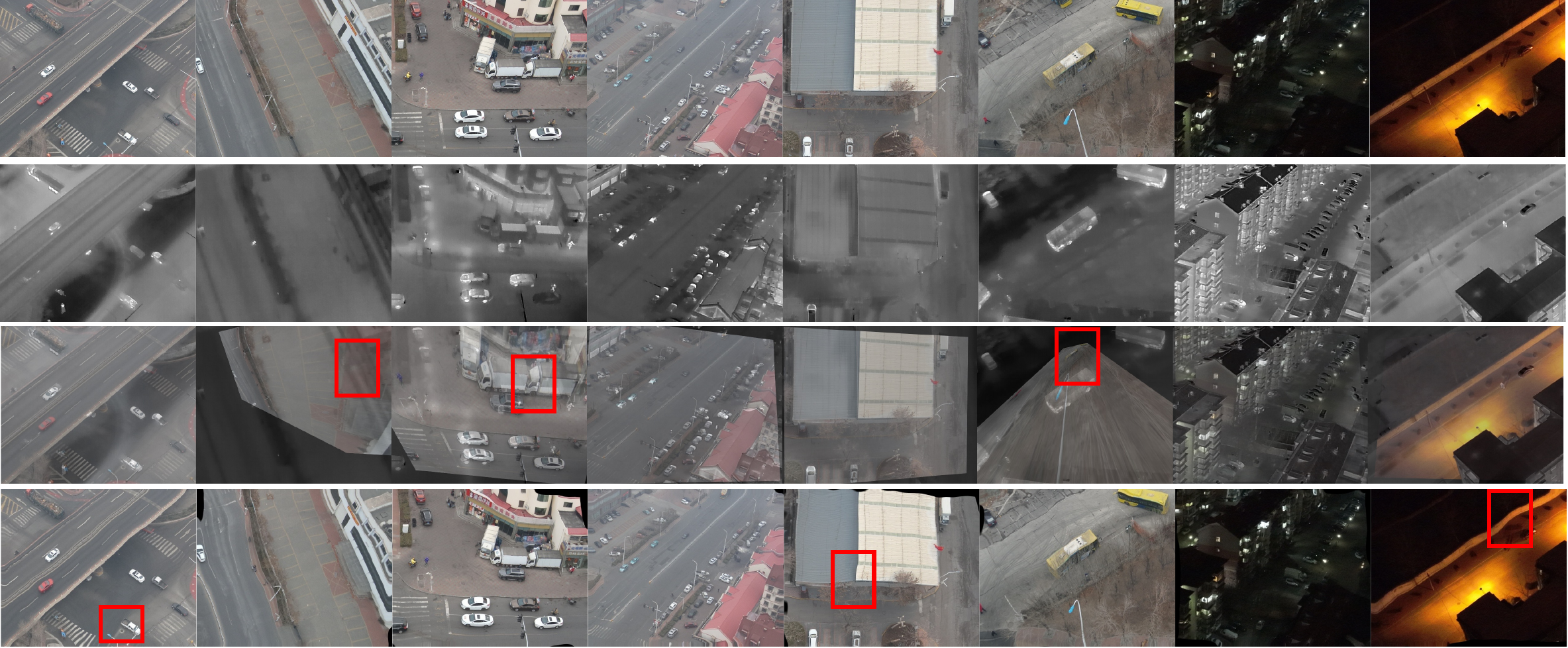} 
    \caption{Comparison of registration results using LoFTR and SuperFusion under different viewpoints and lighting conditions.
The first and second rows present the RGB and TIR channel images, respectively.
The third and fourth rows showcase the registration outcomes of the LoFTR and SuperFusion methods.
Regions with significant registration discrepancies are highlighted.}
    \label{fig:a2}
\end{figure*}

\begin{figure}
    \centering
    \includegraphics[width=1\linewidth]{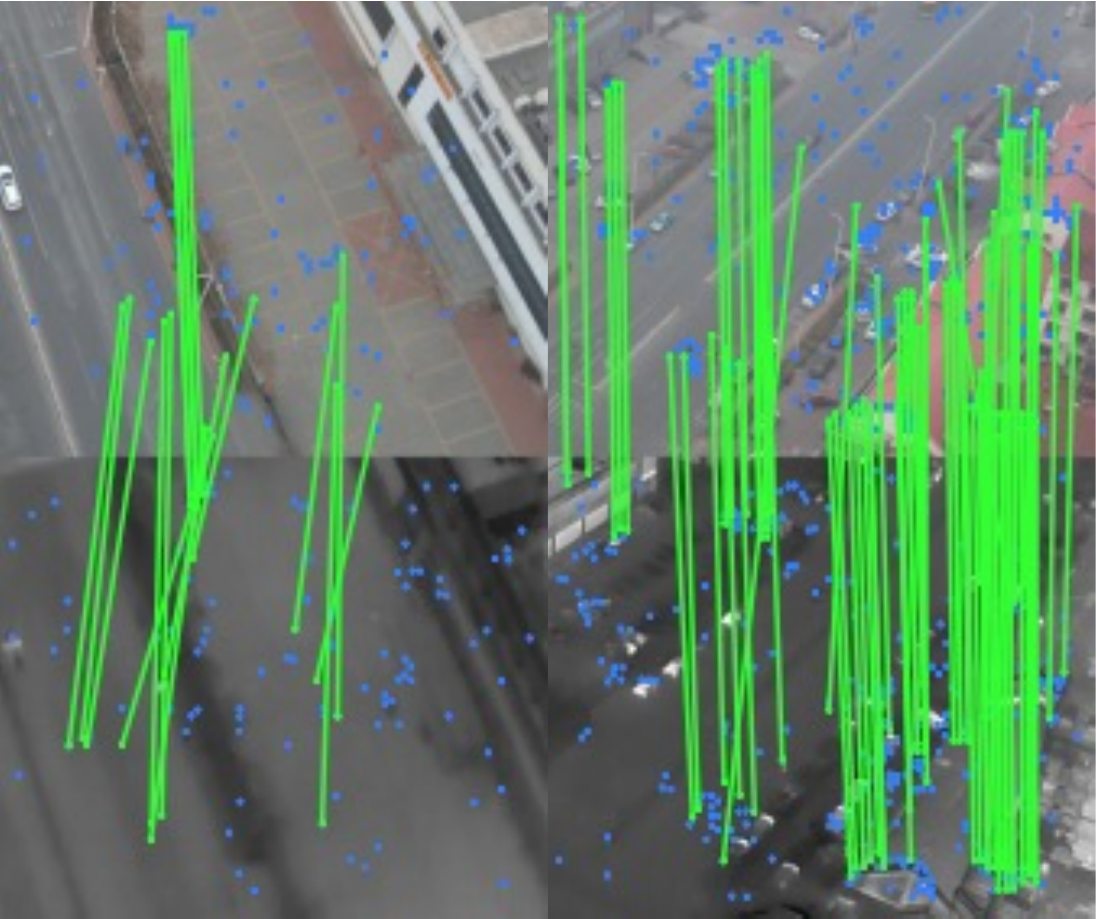}
    \caption{Visualization of intermediate point registration results using the LoFTR method in sparse and dense sample scenarios.}
    \label{fig:a3}
\end{figure}

\noindent\textbf{Experimental observations}

 We utilized Loftr and SuperFusion to register RGB and TIR images separately and experimented by replacing the original RGB or TIR images with the fused images during model training and testing. The registration results in different scenarios can be observed in Figures \ref{fig:a2} and \ref{fig:a3}. The performance metrics of different registration methods can be found in Figure \ref{fig:alal}. Below are some interesting findings:




\begin{enumerate}[label=(\roman*):,leftmargin=12pt]
    \item \textbf{Observation on Registration Performance.}
    The experimental results demonstrate that Loftr and SuperFusion exhibit distinct advantages and characteristics in generating fused RGB images. Loftr focuses on precise feature matching and geometric alignment, ensuring that the fused RGB image is spatially well-aligned with the TIR image, with each pixel accurately corresponding to its counterpart. As shown in Figure \ref{fig:a2}, Loftr performs well in images with high sample density, displaying strong spatial stability—likely due to the greater availability of feature mapping information provided by the dense samples. However, its performance deteriorates in sparser scenes, sometimes leading to issues such as ghosting and overlapping artifacts, making it challenging to proceed with subsequent detection steps.

    In contrast, SuperFusion excels at handling sparse scenes where Loftr struggles, effectively preserving sample information and image features. However, it may impact the geometric characteristics of certain scenes, such as the vertical structures of bridges, whereas Loftr remains largely unaffected in such scenarios.

\begin{figure*}[h]
    \centering
        \begin{minipage}{\linewidth}
            \centering
            \begin{minipage}{0.49\linewidth}
                \centering
                \includegraphics[width=\linewidth]{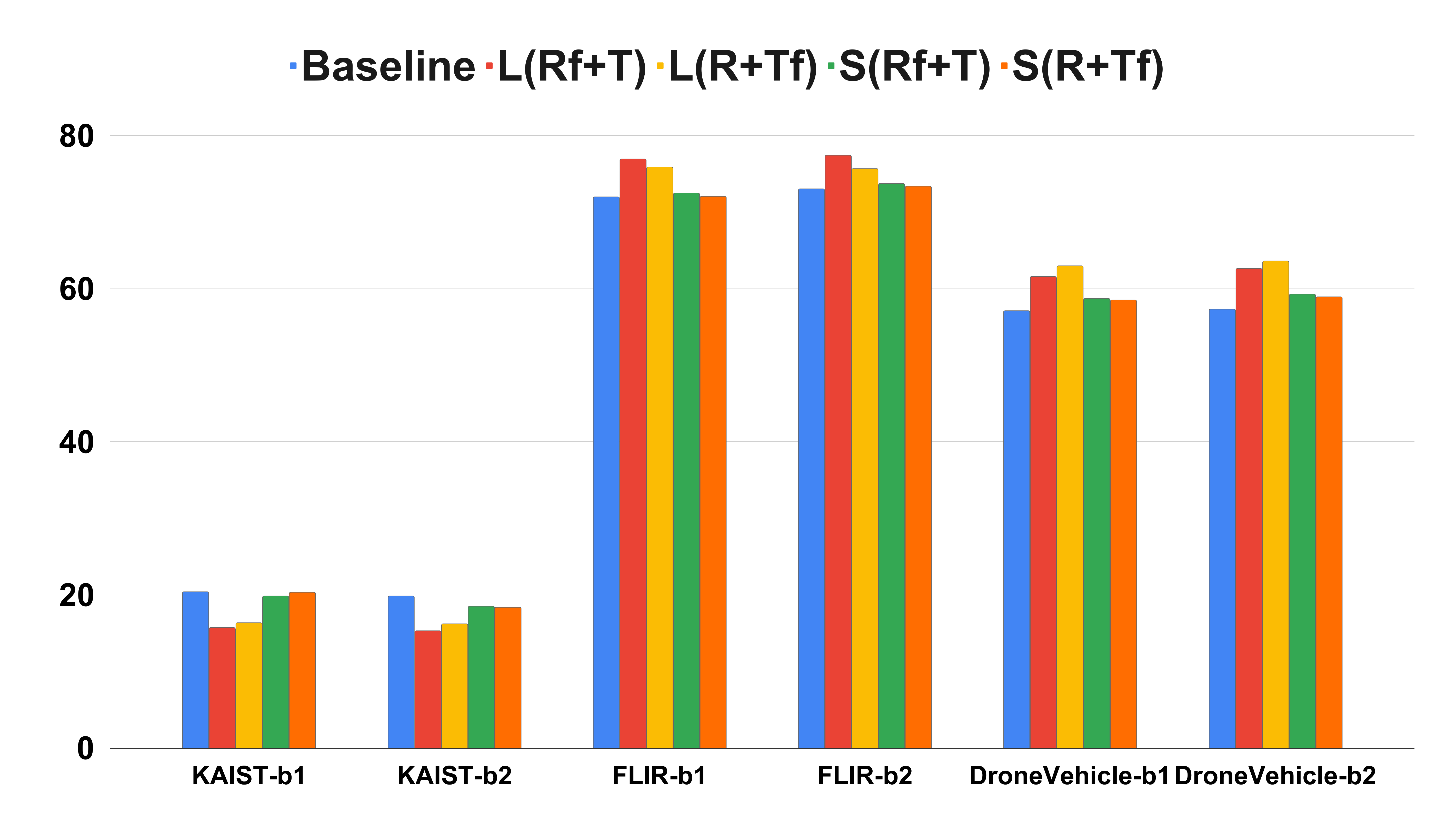} 
            \end{minipage}%
            \hfill
            \begin{minipage}{0.49\linewidth}
                \centering
                \includegraphics[width=\linewidth]{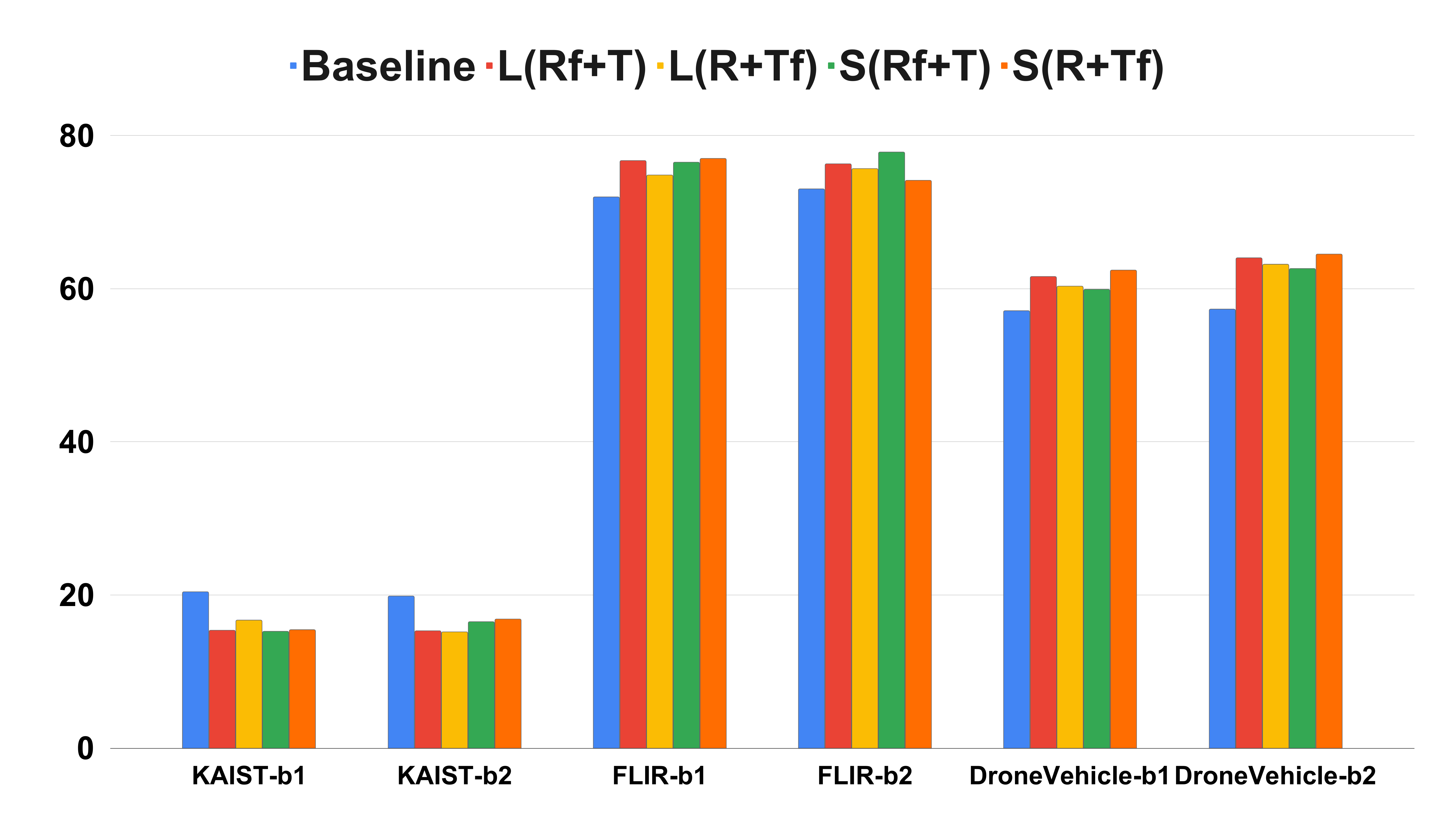} 
            \end{minipage}
        \end{minipage}
        \vspace{-1.0em}
    \caption{The performance metrics of different registration methods at various stages are presented. The image on the left represents registration during the training phase, while the image on the right represents registration during the testing phase. In this figure, b1 corresponds to the method where the registered image replaces the original RGB image, and b2 corresponds to the method where the registered image replaces the original TIR image.}
    \label{fig:alal}
\end{figure*}

\item \textbf{Observations on Registration Methods.} 
The results in Figure \ref{fig:alal} indicate that training the multispectral object detection model with Loftr-registered data yields a substantial increase in recognition accuracy, whereas training with SuperFusion-processed data shows limited impact. During testing, however, both Loftr and SuperFusion enhance recognition accuracy. This advantage is likely due to Loftr's ability to address data inconsistencies via feature alignment during training, thereby improving data quality and facilitating more effective feature learning.

While SuperFusion is effective for multimodal fusion, it may introduce redundancy and complexity in the training data, potentially diverting the model’s focus from key features and limiting accuracy gains. In testing, both methods improve recognition accuracy by refining data quality or enriching feature information. Importantly, both registration frameworks perform best when generating RGB data based on the TIR reference, likely because the TIR-based RGB retains essential thermal information, supporting reliable performance in challenging conditions such as low light, smoke, or nighttime environments.
\item \textbf{Observations on Application Scenarios.} 
Experimental results indicate that Loftr excels in scenarios with significant rotational deviation or displacement between RGB and TIR images. This effectiveness is likely due to Loftr’s precise feature matching and geometric transformations, which effectively mitigate spatial misalignments. Conversely, SuperFusion demonstrates greater suitability in environments affected by adverse weather or low resolution, where it efficiently integrates multimodal data despite these challenges.
\end{enumerate}

\vspace{-0.75em}


\section{Optimal Combination of Individual Techniques}
\begin{figure*}[h]
    \centering
    \begin{minipage}{0.32\linewidth}
        \centering
        \includegraphics[width=\linewidth]{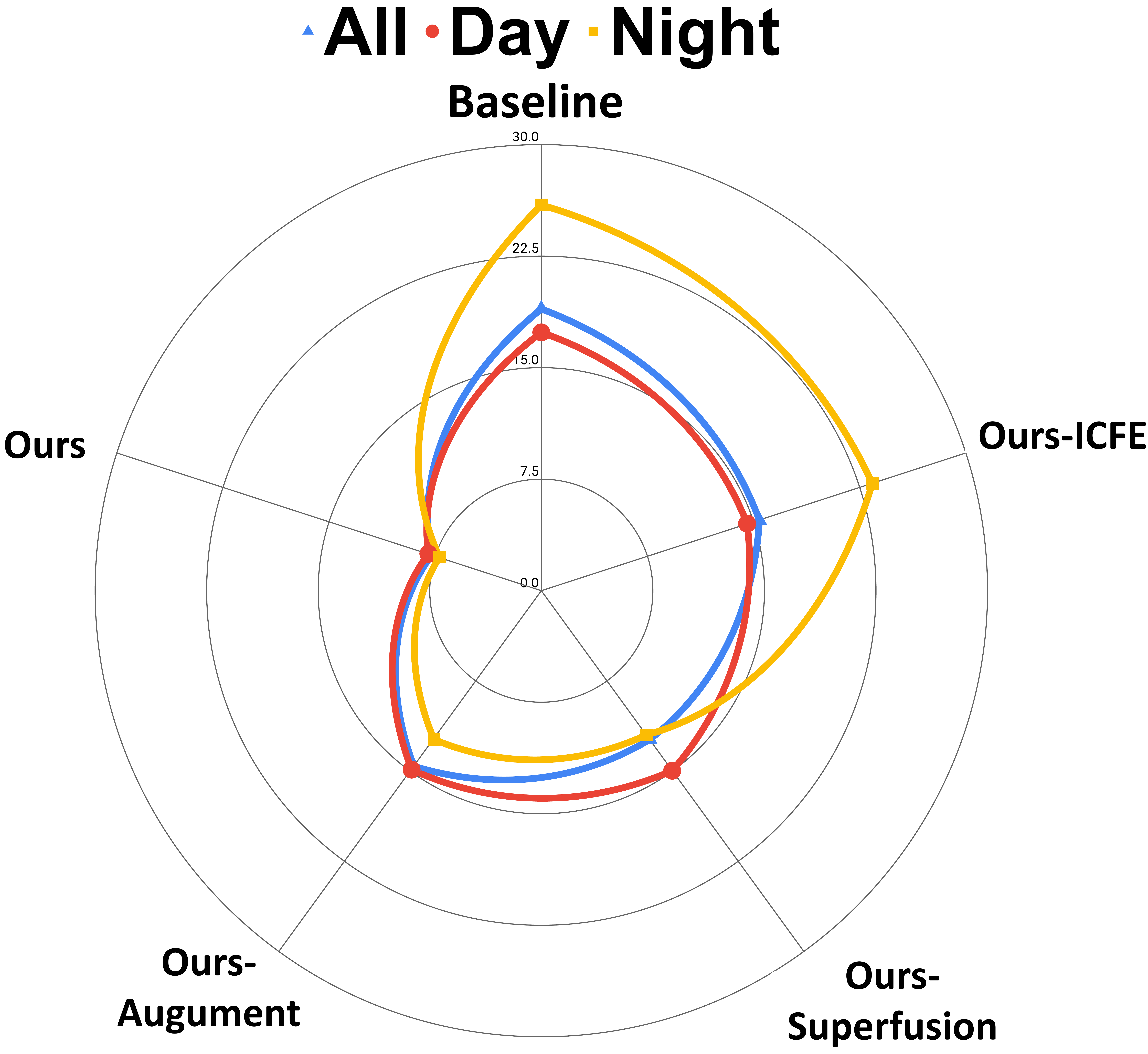}
        \label{fig:enter-label-1}
    \end{minipage}%
    \hfill
    \begin{minipage}{0.32\linewidth}
        \centering
        \includegraphics[width=\linewidth]{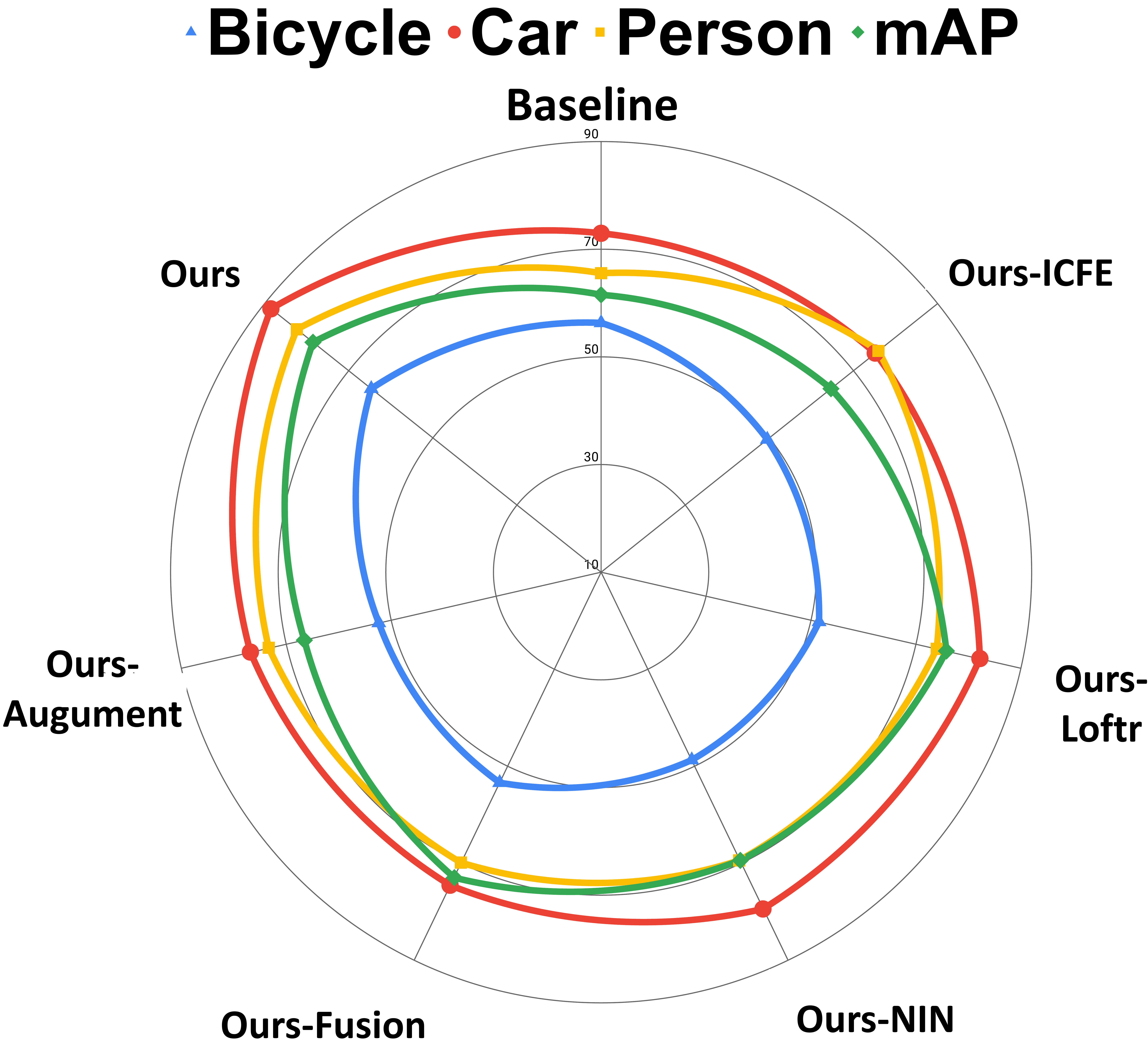}
        \label{fig:enter-label-2}
    \end{minipage}%
    \hfill
    \begin{minipage}{0.32\linewidth}
        \centering
        \includegraphics[width=\linewidth]{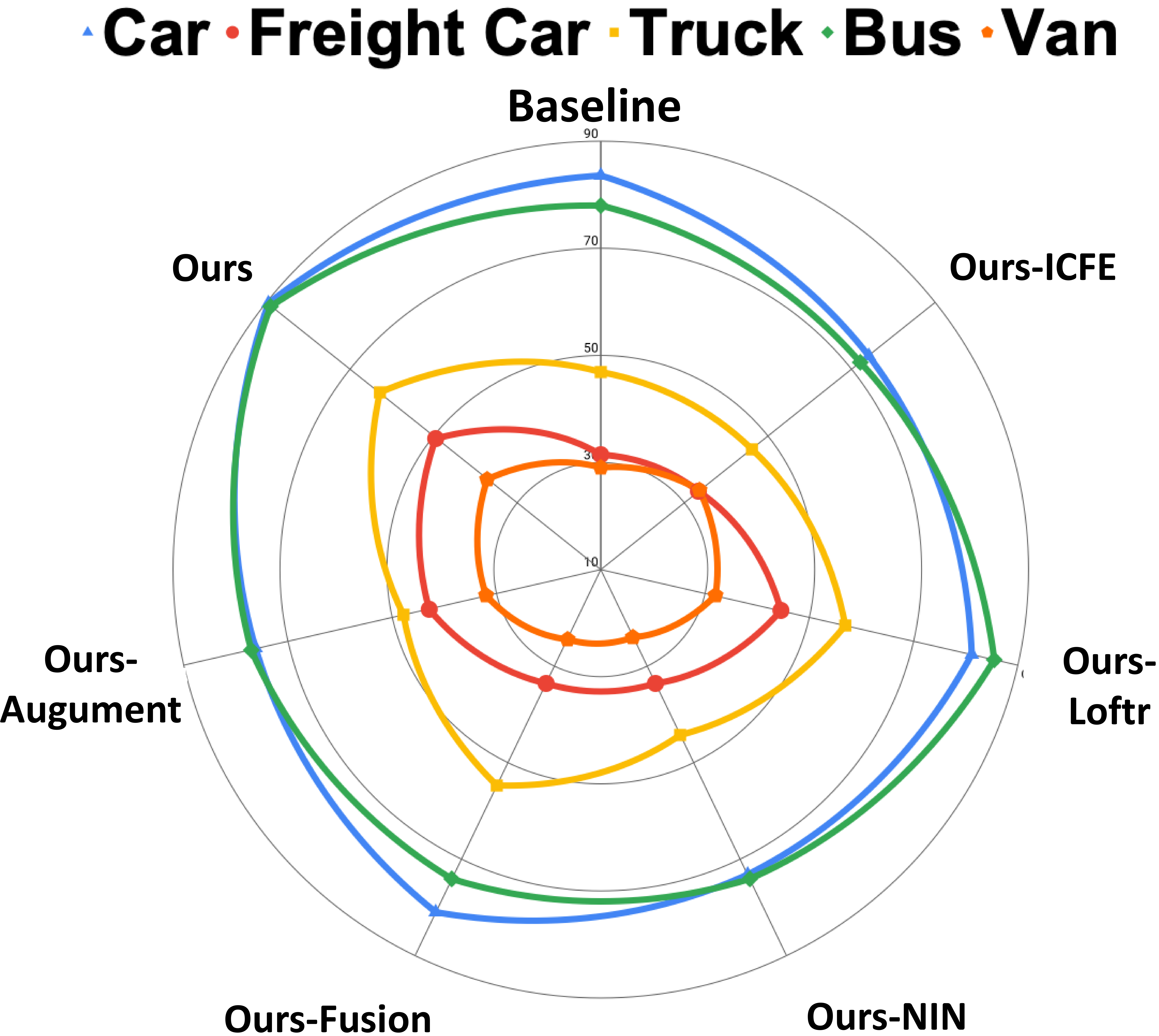}
        \label{fig:enter-label-3}
    \end{minipage}
    \vspace{-1.2em}
    \caption{Ablation experiment results on the KAIST, FLIR, and DroneVehicle datasets. The experimental configurations strictly adhere to the setups outlined in the ``Best Technique Combination''.}
    \vspace{-1.2em}
    \label{fig:ab}
\end{figure*}

In the previous section, we evaluated various training techniques for multispectral object detection under consistent conditions. However, extending a single-modality model to dual-modality with only one technique often yields suboptimal performance, as no single method fully addresses challenges like feature misalignment, overfitting, and fusion conflicts. Therefore, given the diverse methods in multispectral frameworks, relying on a single technique to enhance model performance is impractical. Our benchmark analysis highlights effective combinations of techniques and offers new insights for designing multispectral object detection models.

\begin{tcolorbox}[colback=black!5!white, colframe=black!75!black, title=Best Technique Combinations]
With the optimal hyperparameter settings, the following combinations are recommended:

\textbf{KAIST Dataset:} A single-modality Co-Detr-based model, utilizing the Vit-L backbone and ICFE feature fusion, applies a dual-channel synchronization enhancement strategy through Stitcher, multi-scale scaling, and illumination augmentation. The model leverages SuperFusion for alignment within the test set, resulting in significant improvements in detection accuracy.

\textbf{FLIR Dataset:} A single-modality Co-Detr-based model, integrating the Vit-L backbone with ICFE+NIN feature fusion, achieves dual-channel synchronization enhancement through FastMosaic, multi-scale scaling, and illumination augmentation. The model employs LoFTR for alignment on the test set, delivering exceptional performance.

\textbf{DroneVehicle Dataset:} A single-modality Co-Detr-based model, combining the Vit-L backbone and ICFE+NIN feature fusion, applies Loftr for alignment within the test set. The model adopts complementary enhancement strategies (CLAHE for the RGB channel and random illumination and contrast enhancement for the TIR channel) and applies synchronization augment for both channels using Stitcher and multi-scale scaling. This approach leads to significant improvements in detection performance under low-light conditions.
\end{tcolorbox}

\subsection{Optimal Trick Combinations and Ablation Study}

We have summarized the optimal technique combinations for the KAIST, FLIR, and DroneVehicle datasets above. Additionally, we conducted detailed ablation studies to validate the effectiveness of these combinations, as shown in Figure \ref{fig:ab}. For each dataset, we tested 5 to 6 different combination variants by removing or substituting certain techniques. The results consistently demonstrate the significant effectiveness of our selected combinations, and the observed performance variations on specific samples are highly consistent with the conclusions we presented in Sections 3.2, 3.3, and 3.4.

\subsection{Comparison with Leading Frameworks}

To further validate the effectiveness of the optimized single-modality model based on the best technique combinations, we compared it with other advanced frameworks specifically designed for multispectral object detection, including MBNet, MLPD, and MSDS-RCNN. As shown in Tables \ref{tab:b1}, \ref{tab:b2}, \ref{tab:b3}, by organically integrating our training techniques into the single-modality model, the optimized model consistently outperforms previously well-designed multispectral detection frameworks on both small-scale and large-scale datasets. 

\subsection{Transferring Technique Combinations}

The final plausibility check is to determine whether certain technique combinations remain effective across multiple multispectral object detection datasets. To this end, we selected the combination of \textbf{``Loftr for test alignment + ICFE for feature fusion''}, as these two techniques consistently demonstrated optimal performance in the majority of scenarios covered in Sections 3.2, 3.3, and 3.4. This combination also performed comparably to other top-performing combinations on the FLIR and DroneVehicle datasets. Specifically, we evaluated this approach on two additional open-source multispectral detection datasets: (i) the LLVIP dataset, (ii) the CVC-14 dataset. In these transfer studies, we strictly adhered to the ``best configuration point'' settings outlined in Section 3.1.

\begin{table}[h] \footnotesize
\centering
\renewcommand{\arraystretch}{0.8} 
\setlength{\tabcolsep}{5pt} 
\captionsetup{justification=centering, labelsep=period, font=bf} 
\caption{\footnotesize{Performance metrics of models with and without our strategy on the LLVIP and CVC-14 datasets. The results are averaged over multiple independent runs, with the standard deviations provided.}} 
\vspace{-0.7em}
\label{tab:trans}
\begin{tabular}{@{}lcccc@{}}
\toprule
\textbf{Method} & \textbf{Strategy} & \multicolumn{2}{c}{\textbf{LLVIP}} & \textbf{CVC-14} \\ 
\cmidrule(lr){3-4} \cmidrule(lr){5-5}
 &  & \textbf{mAP50(\%)} & \textbf{mAP(\%)} & \textbf{\boldmath $MR^2$ (\%)↓}

 \\ 
\midrule
SSD \cite{Liu_2016}              & w/o & $90.25_{\pm 1.76}$ & $53.52_{\pm 2.45}$ & $68.39_{\pm 1.78}$ \\
                 & with    & $92.13_{\pm 2.45}$ & $54.39_{\pm 2.31}$ & $37.16_{\pm 2.18}$ \\
RetinaNet \cite{8237586}        & w/o & $94.81_{\pm 2.13}$ & $55.18_{\pm 1.29}$ & $47.87_{\pm 2.75}$ \\
                 & with    & $95.15_{\pm 1.89}$ & $57.87_{\pm 2.48}$ & $29.63_{\pm 1.32}$ \\
Cascade R-CNN \cite{10055028}    & w/o & $95.12_{\pm 2.23}$ & $56.81_{\pm 2.61}$ & $42.36_{\pm 2.91}$ \\
                 & with    & $95.58_{\pm 1.68}$ & $57.99_{\pm 1.35}$ & $22.15_{\pm 1.54}$ \\
Faster R-CNN \cite{10055028}    & w/o & $94.63_{\pm 2.78}$ & $54.53_{\pm 2.43}$ & $51.97_{\pm 1.97}$ \\
                 & with    & $94.97_{\pm 2.11}$ & $56.15_{\pm 1.95}$ & $24.31_{\pm 1.87}$ \\
DDQ-DETR \cite{zhang2023densedistinctqueryendtoend}         & w/o & $93.91_{\pm 1.67}$ & $58.67_{\pm 1.49}$ & $52.78_{\pm 2.41}$ \\
                 & with    & $94.86_{\pm 2.26}$ & $60.13_{\pm 1.87}$ & $26.51_{\pm 1.53}$ \\
\bottomrule
\end{tabular}
\end{table}

\begin{table}[h]
\footnotesize
\centering
\renewcommand{\arraystretch}{1}  
\setlength{\tabcolsep}{16pt} 
\captionsetup{justification=centering, labelsep=period, font=bf} 
\caption{\footnotesize{Comparison of our most effective detection model with other advanced frameworks on the KAIST dataset. We use bold red font and underline to highlight the best results.}} 
\vspace{-1.0em}
\begin{tabular}{@{}lcccc@{}}
\toprule
\multirow{2}{*}{\textbf{Method}} & \multicolumn{3}{c}{\textbf{\boldmath $MR^2$ (\%)↓}}
 \\ 
\cmidrule(lr){2-4}
 & \textbf{All} & \textbf{Day} & \textbf{Night} \\ 
\midrule
FusionRPN+BF \cite{Knig2017FullyCR}        & 18.31 & 19.54 & 16.33 \\
IAF-RCNN \cite{li2018illuminationawarefasterrcnnrobust}            & 15.55 & 14.97 & 16.89 \\
IATDNN-IAMSS \cite{guan2018fusionmultispectraldatailluminationaware}        & 14.41 & 14.30 & 15.29 \\
MBNet \cite{zhou2020improvingmultispectralpedestriandetection}               & 8.43  & 8.79  & 8.10  \\
MLPD \cite{9496129}                & 7.21  & \textbf{\underline{\textcolor{red}{6.83}}} & 7.68  \\
MSDS-RCNN \cite{li2018multispectralpedestriandetectionsimultaneous}           & 7.34  & 8.98  & 6.94  \\
\textbf{Ours}       & \textbf{\underline{\textcolor{red}{6.23}}} & 6.91 & \textbf{\underline{\textcolor{red}{6.19}}} \\
\bottomrule
\end{tabular}
\label{tab:b1}
\end{table}

\begin{table}[h]
\footnotesize
\centering
\renewcommand{\arraystretch}{1} 
\setlength{\tabcolsep}{6pt} 
\captionsetup{justification=centering, labelsep=period, font=bf} 
\caption{\footnotesize{Comparison of our most effective detection model with other advanced frameworks on the FLIR dataset. We use bold red font and underline to highlight the best results.}}
\vspace{-1.0em}
\begin{tabular}{@{}lcccc@{}}
\toprule
\multirow{2}{*}{\textbf{Method}} & \multicolumn{3}{c}{\textbf{AP50 (\%)}} & \multirow{2}{*}{\textbf{mAP (\%)}} \\ 
\cmidrule(lr){2-4}
 & \textbf{Bicycle} & \textbf{Car} & \textbf{Person} & \\ 
\midrule
MMTOD-CG \cite{9025398}            & 50.38 & 70.61 & 63.42 & 61.47 \\
MMTOD-UNIT \cite{9025398}          & 49.28 & 70.78 & 64.33 & 61.46 \\
CFR \cite{9191080}                 & 57.95 & 84.92 & 74.46 & 72.44 \\
BU-ATT \cite{10.1145/3418213}              & 56.01 & 87.11 & 76.08 & 73.06 \\
BU-LTT \cite{10.1145/3418213}              & 57.43 & 86.31 & 75.65 & 73.13 \\
CFT \cite{qingyun2022crossmodalityfusiontransformermultispectral}                 & 61.44 & \textbf{\underline{\textcolor{red}{89.55}}} & 84.28 & 78.42 \\
\textbf{Ours}       & \textbf{\underline{\textcolor{red}{68.71}}} & 89.51 & \textbf{\underline{\textcolor{red}{85.30}}} & \textbf{\underline{\textcolor{red}{81.17}}} \\
\bottomrule
\end{tabular}
\label{tab:b2}
\end{table}

\begin{table}[h]
\footnotesize
\centering
\renewcommand{\arraystretch}{1} 
\setlength{\tabcolsep}{3pt} 
\captionsetup{justification=centering, labelsep=period, font=bf} 
\caption{\footnotesize{Comparison of our most effective detection model with other advanced frameworks on the DroneVehicle dataset. We use bold red font and underline to highlight the best results.}} 
\vspace{-1.0em}
\begin{tabular}{@{}lc@{\hspace{3pt}}cccc@{\hspace{3pt}}c@{}}
\toprule
\multirow{2}{*}{\textbf{Method}} & \multicolumn{5}{c}{\textbf{AP50 (\%)}} & \multirow{2}{*}{\textbf{mAP (\%)}} \\ 
\cmidrule(lr){2-6}
 & \textbf{Car} & \textbf{Freight Car} & \textbf{Truck} & \textbf{Bus} & \textbf{Van} & \\ 
\midrule
RetinaNet-OBB \cite{8237586}          & 65.36 & 15.69 & 32.81 & 61.34 & 16.26 & 38.29 \\
Mask R-CNN \cite{10055028}             & 88.98 & 36.84 & 47.79 & 78.17 & 36.65 & 57.69 \\
Cascade Mask R-CNN \cite{10055028}     & 80.95 & 31.00 & 38.27 & 66.62 & 25.01 & 48.37 \\
UA-CMDet \cite{9759286}               & 87.35 & 41.27 & 62.69 & 84.17 & 39.82 & 63.06 \\
CALNet \cite{10.1145/3581783.3612651}                 & 86.32 & 60.67 & 67.15 & 86.52 & 53.68 & 70.87 \\
TSFADet \cite{yuan2022translationscalerotationcrossmodal}                & 89.01 & 51.97 & 68.51 & 83.06 & 46.95 & 67.9 \\
Gliding Vertex \cite{Xu_2021}         & 89.99 & 42.75 & 59.71 & 79.79 & 44.19 & 63.29 \\
\textbf{Ours}          & \textbf{\underline{\textcolor{red}{92.05}}} & \textbf{\underline{\textcolor{red}{63.39}}} & \textbf{\underline{\textcolor{red}{71.95}}} & \textbf{\underline{\textcolor{red}{88.93}}} & \textbf{\underline{\textcolor{red}{57.12}}} & \textbf{\underline{\textcolor{red}{74.69}}} \\
\bottomrule
\end{tabular}
\label{tab:b3}
\end{table}

As shown in Table \ref{tab:trans}:
the selected technique combination significantly improved the performance of the single-modality model on various multispectral datasets in most cases, particularly in scenarios with complex backgrounds and varying lighting conditions. This combination consistently enhanced model performance across different datasets, with the CVC-14 dataset showing a maximum accuracy improvement of over 31.23\%. The strong transferability of this technique combination suggests its potential to serve as a robust baseline for future research in multispectral object detection, while also offering new training strategies for optimizing single-modality detection models.

\section{Conclusion}

Multispectral object detection is a rapidly advancing field, yet significant challenges remain in effectively integrating multimodal information to adapt to diverse environmental conditions. In this study, we propose a standardized benchmark with fair and consistent experimental setups to drive progress in this domain. We conducted extensive experiments across multiple public datasets, focusing on three critical aspects of multispectral detection: multimodal feature fusion, dual-modality data augmentation, and registration alignment. Through a comprehensive analysis of our results, we identified the most effective technique combinations and established new performance benchmarks for multispectral object detection.

Additionally, we introduce a novel training strategy to optimize single-modality models for dual-modality tasks, laying the groundwork for adapting high-performing single-modality models to dual-modality scenarios. We believe that the strong baselines and optimized technique combinations presented in this work will facilitate fairer and more practical evaluations in multispectral object detection research. This work sets a robust foundation for future studies and opens new avenues for enhancing multispectral object detection performance.


%






\ifCLASSOPTIONcaptionsoff
  \newpage
\fi



%
\bibliographystyle{IEEEtran}
\bibliography{references}
%

\begin{IEEEbiography}[{\includegraphics[width=1in,height=1.25in,clip,keepaspectratio]{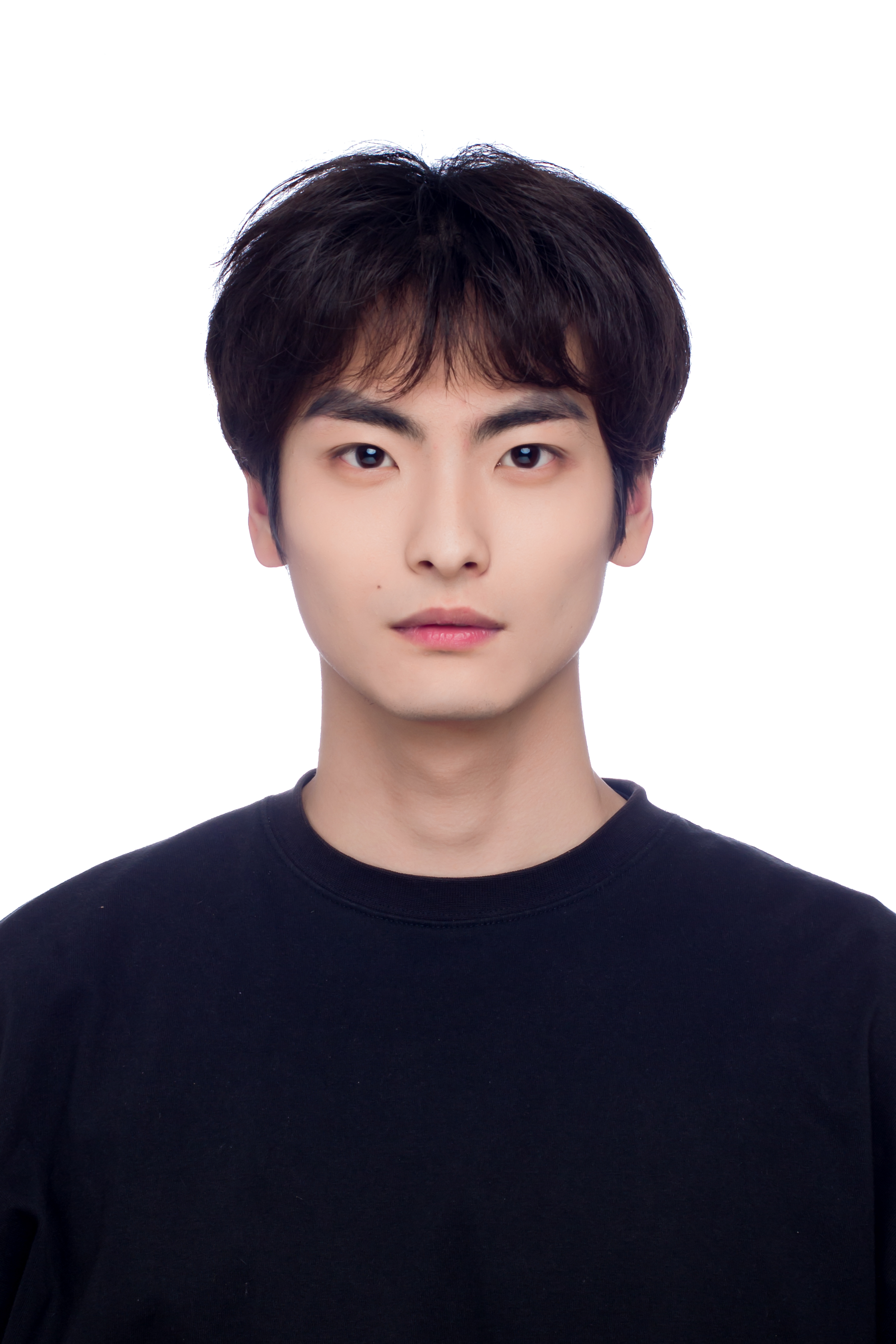}}]{Chen Zhou} is a master's student at Beijing Forestry University and currently works at the TeleAI Artificial Intelligence Research Institute of China Telecom. His research focuses on computer vision, speech-driven lip motion, and multimodal generative algorithms. He has won several prestigious awards, including the championship in the Global AI Technology Innovation Competition for Dual-Spectrum Object Detection in Drone Perspectives, the championship in the Ruikang Robotics Developer Competition for Algorithm Optimization, and third place in the "Smart Balance House" AI Challenge for the Target Recognition System Speed Competition.
\end{IEEEbiography}

\begin{IEEEbiography}[{\includegraphics[width=1in,height=1.25in,clip,keepaspectratio]{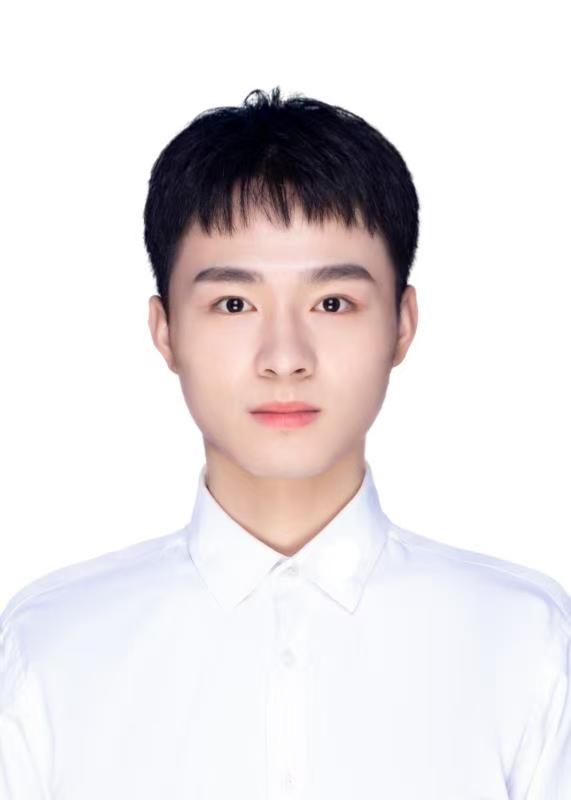}}]{Peng Cheng} is a master's student at Beijing Forestry University. His research focuses on computer vision, data compression, and multimodal technologies. He has achieved remarkable results in various competitions, including the championship in the SEED Jiangsu Big Data Development and Application Competition, the Global AI Technology Innovation Competition for Dual-Spectrum Object Detection in Drone Perspectives, and the CVPR Challenge on Low-light Object Detection. Altogether, he has won over 20 prestigious competition awards.
\end{IEEEbiography}

\begin{IEEEbiography}[{\includegraphics[width=1in,height=1.25in,clip,keepaspectratio]{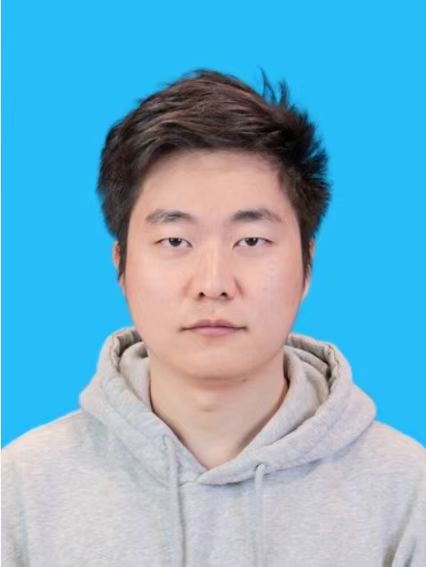}}]{Junfeng Fang} obtained his PhD from the University of Science and Technology of China. His research primarily focuses on Model Editing and LLM Explainability. He has published in top conferences and journals, including NeurIPS, KDD, ICLR, and TKDE. He has also served as a reviewer for major conferences and journals such as ICLR, KDD, NeurIPS, ICML, and TKDE.
\end{IEEEbiography}

\begin{IEEEbiography}[{\includegraphics[width=1in,height=1.25in,clip,keepaspectratio]{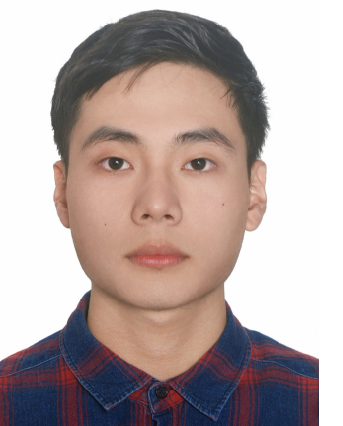}}]{Yibo Yan} is currently a Ph.D. candidate of Artificial Intelligence Thrust, Hong Kong University of Science and Technology (Guangzhou) and Department of Computer Science and Engineering, Hong Kong University of Science and Technology. His primary research interest include multimodal learning, large language model, and natural language processing.
\end{IEEEbiography}

\begin{IEEEbiography}[{\includegraphics[width=1in,height=1.25in,clip,keepaspectratio]{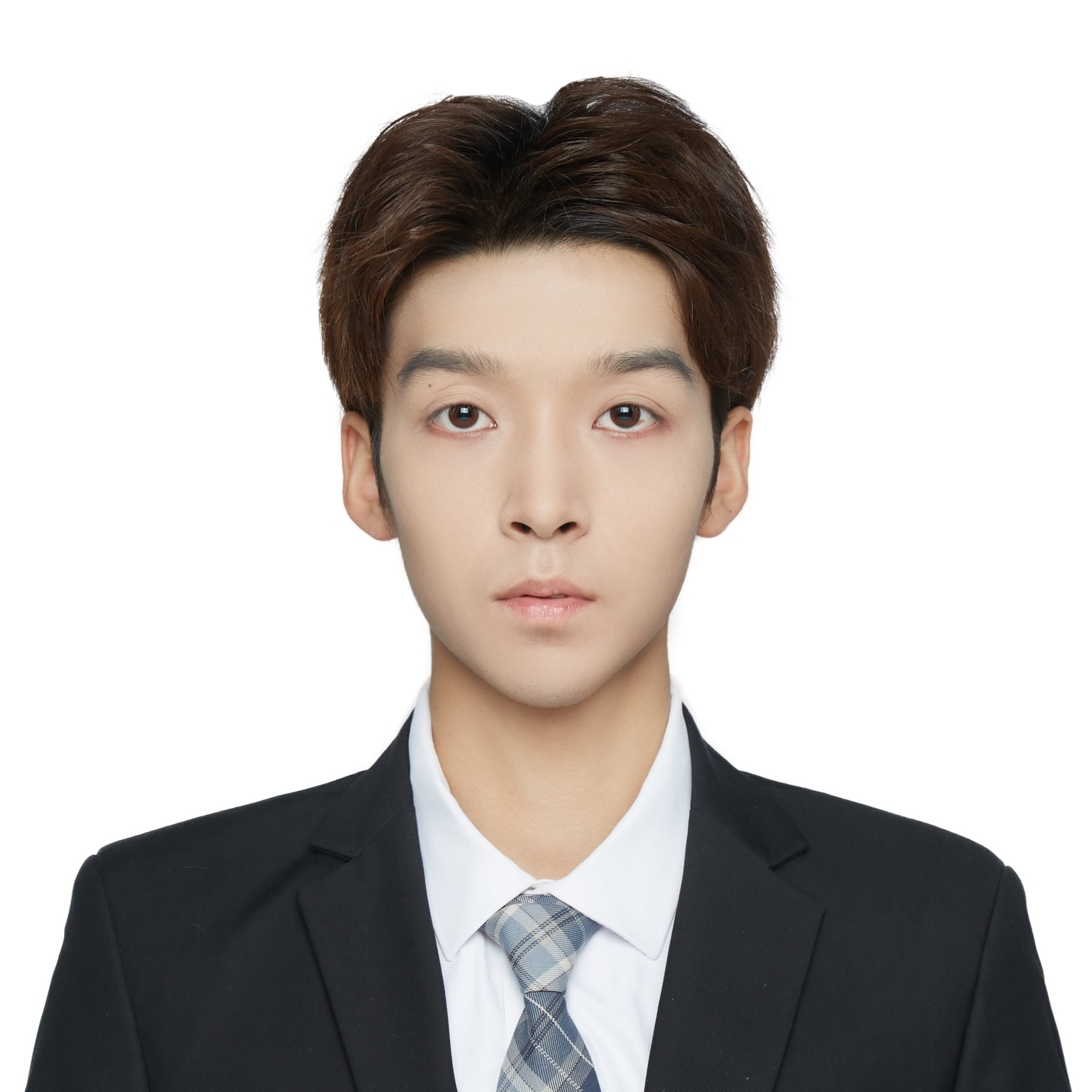}}]{Yifan Zhang} (yifanzhang.cs@gmail.com) is with the State Key Laboratory of Multimodal Artificial Intelligence Systems (MAIS),  Institute of Automation, Chinese Academy of Sciences (CASIA), Beijing 100190, China, and also with the School of Artificial Intelligence, University of Chinese Academy of Sciences (UCAS), Beijing 100049, China. 
\end{IEEEbiography}

\begin{IEEEbiography}[{\includegraphics[width=1in,height=1.25in,clip,keepaspectratio]{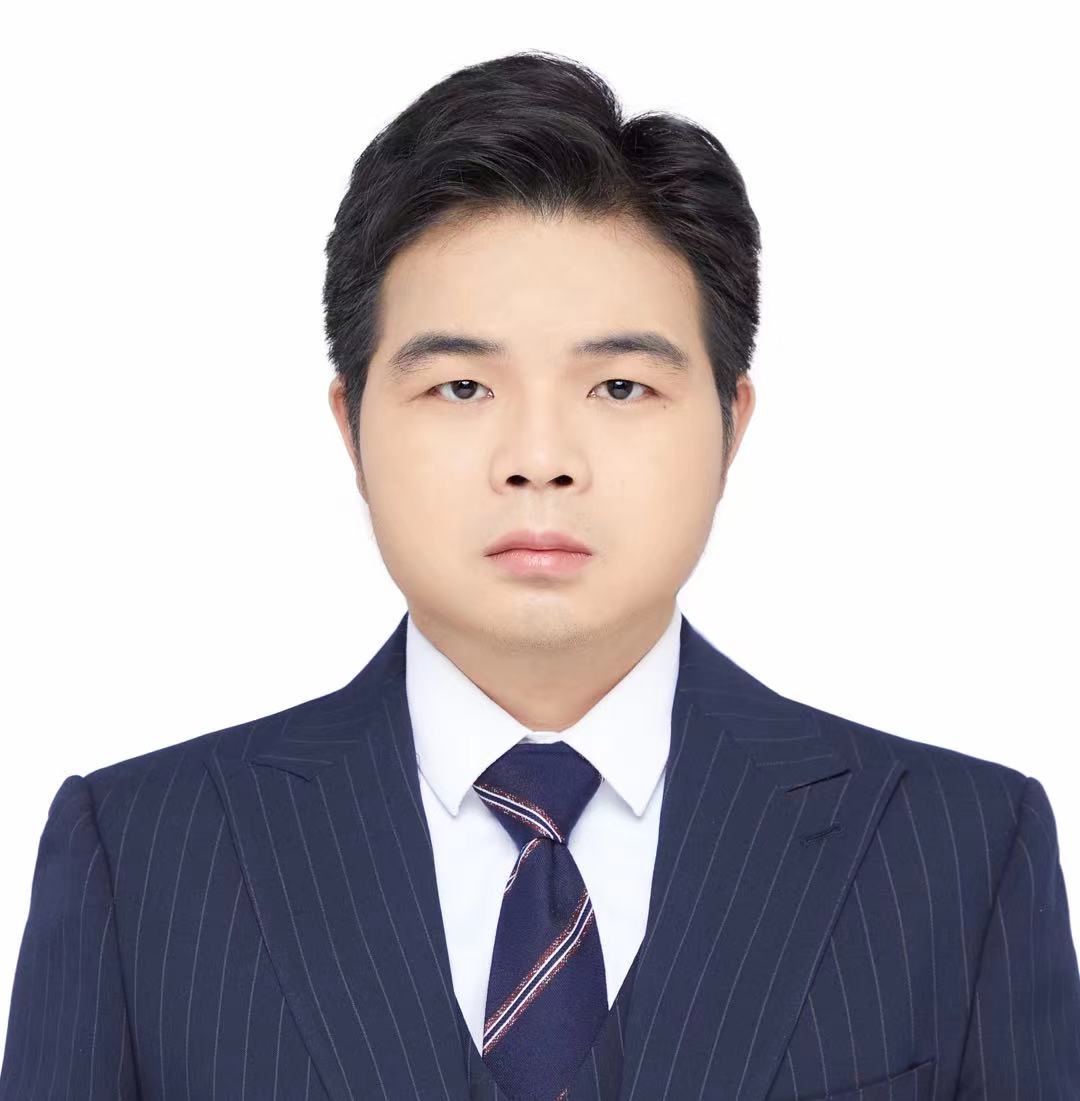}}]{Xiaojun Jia} received his Ph.D. degree in  State Key Laboratory of Information Security, Institute of Information Engineering, Chinese Academy of Sciences and School of Cyber Security, University of Chinese Academy of Sciences, Beijing. He is now a Research Fellow in Cyber Security Research Centre @ NTU, Nanyang Technological University, Singapore. His research interests include computer vision, deep learning and adversarial machine learning.
\end{IEEEbiography}

\begin{IEEEbiography}[{\includegraphics[width=1in,height=1.25in,clip,keepaspectratio]{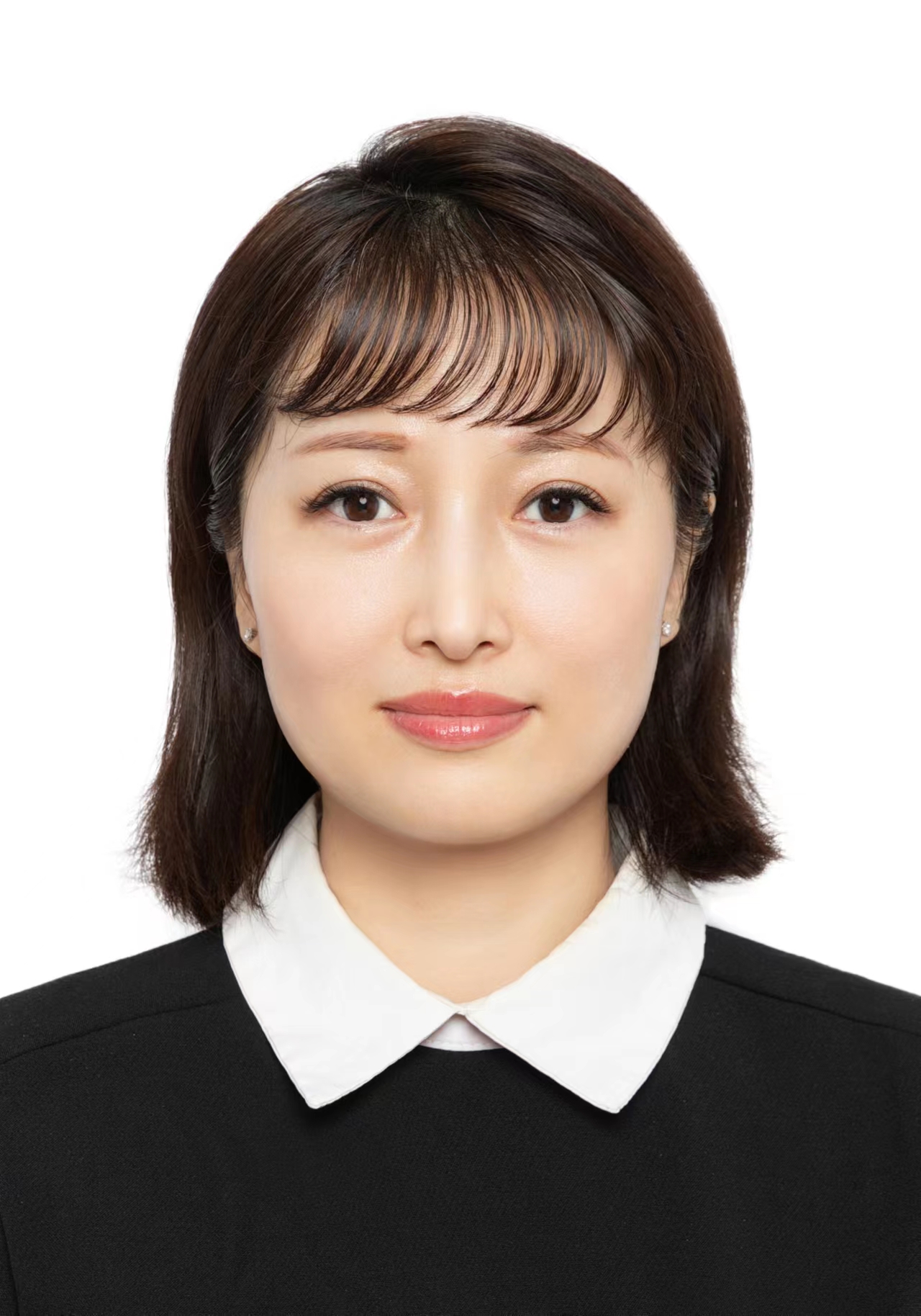}}]{Yanyan Xu} received her Ph.D. from Institute of Software, Chinese Academy of Sciences, and her M.Sc. and B.Sc. from Sun Yat-sen University. She has since joined School of Information Science and Technology, Beijing Forestry University, where she is currently an associate professor. Also, she spent a year as a visiting scholar at Leiden Institute of Advanced Computer Science, Leiden University in the Netherlands. Her research interests include artificial intelligence, speech processing, and large language models.
\end{IEEEbiography}

\begin{IEEEbiography}[{\includegraphics[width=1in,height=1.25in,clip,keepaspectratio]{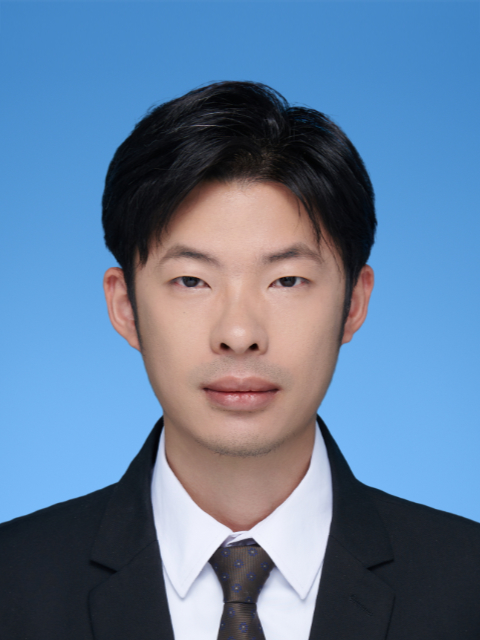}}]{Kun Wang} obtained his PhD from the University of Science and Technology of China and is currently a postdoctoral researcher at Nanyang Technological University. His research primarily focuses on applications of graph structures, covering areas such as sparsification, data mining (especially spatiotemporal forecasting and Earth sciences), and LLM Agent studies. Dr. Wang is dedicated to enhancing the trustworthiness, interpretability, generalization, and robustness of algorithms in deep learning and artificial intelligence. As the first or corresponding author, he has published over 20 papers in top conferences and journals, including TPAMI, ICML, NeurIPS, KDD, ICLR, AAAI, WWW, and TKDE. He has also served as a reviewer for major conferences and journals such as ICLR, KDD, NeurIPS, ICML, and TKDE.
\end{IEEEbiography}

\begin{IEEEbiography}[{\includegraphics[width=1in,height=1.25in,clip,keepaspectratio]{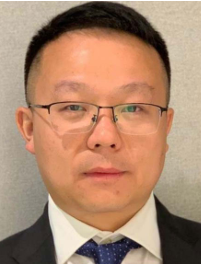}}]{Xiaochun Cao}  (Senior Member, IEEE) received the B.E. and M.E. degrees in computer science from Beihang University, China, and the Ph.D. degree in computer science from the University of Central Florida, Orlando, USA. After graduation, he spent about three years at ObjectVideo Inc., as a Research Scientist. From 2008 to 2012, he was a Professor with Tianjin University, Tianjin, China. He has been a Professor with the Institute of Information Engineering, Chinese Academy of Sciences, since 2012. He is currently with the School of Cyber Security, Sun Yat-sen University, China. He is on the Editorial Boards of IEEE TRANSACTIONS ON IMAGE PROCESSING, IEEE TRANSACTIONS ON MULTIMEDIA, and IEEE TRANSACTIONS ON CIRCUITS AND SYSTEMS FOR VIDEO TECHNOLOGY. From 2004 to 2010, he was a recipient of the Piero Zamperoni Best Student Paper Award from the International Conference on Pattern Recognition.
\end{IEEEbiography}




\newpage
\end{document}


%
\title{DeepGCNs:\\ Making GCNs Go as Deep as CNNs\\ -- Appendix --}
%
%
%
%

\author{Guohao Li\thanks{*equal contribution}\textsuperscript{*}\quad Matthias M\"uller\textsuperscript{*} \quad Guocheng Qian\textsuperscript{*} \quad Itzel C. Delgadillo \quad Abdulellah  Abualshour \quad \\ Ali Thabet\quad Bernard Ghanem\\
		Visual Computing Center,~ KAUST,~ Thuwal,~ Saudi Arabia\\
		{\tt\footnotesize \{guohao.li, matthias.mueller.2, guocheng.qian, itzel.delgadilloperez, abdulellah.abualshour, ali.thabet, bernard.ghanem\}@kaust.edu.sa}}

\markboth{IEEE TPAMI, Special Issue on Graphs in Vision and Pattern Analysis.}
{Shell \MakeLowercase{\textit{et al.}}: Bare Demo of IEEEtran.cls for Computer Society Journals}
\maketitle

\IEEEdisplaynontitleabstractindextext

%
\IEEEpeerreviewmaketitle

\ifCLASSOPTIONcaptionsoff
  \newpage
\fi




\bibliographystyle{IEEEtran}
\bibliography{references}
